\documentclass[runningheads]{llncs}
\usepackage{graphicx}

\usepackage{tikz}
\usepackage{comment}
\usepackage{amsmath,amssymb} 
\usepackage{color}
\usepackage{graphicx}
\usepackage{booktabs}
\usepackage{overpic}
\usepackage{soul}
\usepackage{float}
\usepackage{adjustbox}
\usepackage{wrapfig}
\usepackage{abstract}
\usepackage{xcolor}
\usepackage{subcaption}

\newcommand{\sagie}[1]{{\color{red}[\textbf{Sagie:} #1]}}

\newcommand{\transpose}{\mathrm{T}}

\usepackage[colorlinks,citecolor=green,urlcolor=blue,bookmarks=false,hypertexnames=true]{hyperref}

\usepackage[capitalize]{cleveref}
\crefname{section}{Sec.}{Secs.}
\Crefname{section}{Section}{Sections}
\Crefname{table}{Table}{Tables}
\crefname{table}{Tab.}{Tabs.}

\usepackage[accsupp]{axessibility}  

\usepackage[width=122mm,left=12mm,paperwidth=146mm,height=193mm,top=12mm,paperheight=217mm]{geometry}

\begin{document}
\pagestyle{headings}
\mainmatter

\newcommand*\samethanks[1][\value{footnote}]{\footnotemark[#1]}

\title{Volumetric Disentanglement  for \\ 3D Scene Manipulation} 
\author{Sagie Benaim$^{1}$ \quad Frederik Warburg\thanks{Contributed equally.}$^{2}$ \\ Peter Ebert Christensen\samethanks$^{1}$ \quad Serge Belongie$^{1}$}
\institute{$^{1}$University of Copenhagen \quad $^{2}$Technical University of Denmark}
\authorrunning{Benaim et al.}


\titlerunning{Volumetric Disentanglement for 3D Scene Manipulation}

\maketitle

\begin{abstract}

Recently, advances in differential volumetric rendering enabled significant breakthroughs in the photo-realistic and fine-detailed reconstruction of complex 3D scenes, which is key for many virtual reality  applications. However, in the context of augmented reality, one may also wish to effect semantic manipulations or augmentations of objects within a scene. To this end, we propose a volumetric framework for (i) disentangling or separating, the volumetric representation of a given foreground object from the background, and (ii) semantically manipulating the foreground object, as well as the background. Our framework takes as input a set of 2D masks specifying the desired foreground object for training views, together with the associated 2D views and poses, and produces a foreground-background disentanglement that respects the surrounding illumination, reflections, and partial occlusions, which can be applied to both training and novel views. 
Our method enables the separate control of pixel color and depth as well as 3D similarity transformations of both the foreground and background objects. We subsequently demonstrate the applicability of our framework on a number of downstream manipulation tasks including object camouflage, non-negative 3D object inpainting, 3D object translation, 3D object inpainting, and 3D text-based object manipulation. Full results are given in our project webpage at \url{https://sagiebenaim.github.io/volumetric-disentanglement/}

\keywords{3D Object Editing,  Neural Radiance Fields, Disentanglement}

\end{abstract}

\section{Introduction}

\begin{figure}[h]
\begin{center}
    \includegraphics[width=\textwidth]{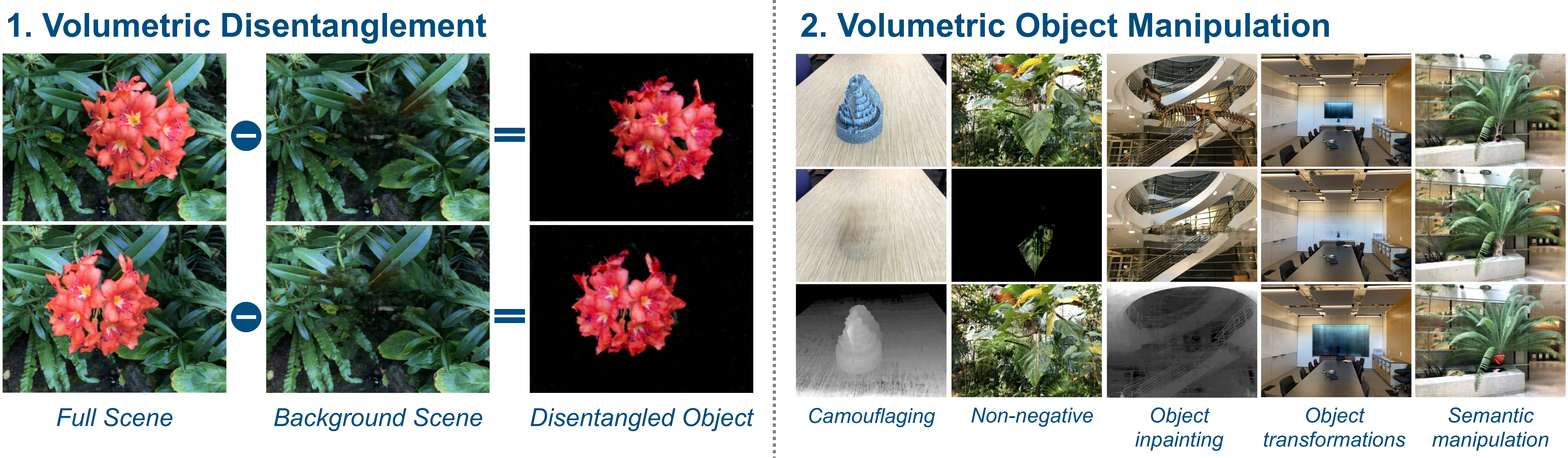} \\ 
    \caption{\textbf{Volumetric disentanglement framework.} We introduce a framework for the volumetric disentanglement of 
    foreground objects as well as the background from a full scene (1). 
    Our volumetric disentanglement can then be used for many downstream tasks of interest to designers and artists in AR applications. We explore some of these tasks in this paper, which includes, as illustrated in (2), the tasks of 3D object camouflage, non-negative 3D inpainting, 3D object inpainting, 3D object transformation, and 3D text-based semantic manipulation.
    }
    \vspace{-0.6cm}
    \label{fig:teaser}
\end{center}
\end{figure}

The ability to interact with a 3D environment is of fundamental importance for many augmented reality (AR) application domains such as interactive visualization, entertainment, games, and robotics~\cite{mekni2014augmented}. Such interactions are often semantic in nature, capturing specified entities in a 3D scene and manipulating them accordingly. To this end, we propose a novel framework for the disentanglement of objects in a 3D scene. Given a small and set of 2D masks delineating a desired foreground object, together with the associated 2D views and poses, our method produces a volumetric representation of both the foreground object and the background. Our volumetric representation enables separate control of pixel color and depth, as well as scale, rotation, and translation of the foreground object and the background. Using this disentangled representation, we demonstrate a suite of downstream manipulation tasks involving both the foreground and background volumes. \cref{fig:teaser} illustrates our proposed volumetric disentanglement and a sampling of the downstream volumetric manipulations that this disentanglement enables. 
We note that while the \emph{foreground/background} terminology is useful for painting a mental picture, we wish to emphasise that the disentanglement is not limited to foreground objects, and works equally well for objects positioned further back (and partially occluded) in a scene.

Neural Radiance Fields (NeRF)~\cite{mildenhall2020representing} recently delivered a significant breakthrough in the ability to reconstruct complex 3D scenes with high fidelity and a high level of detail. At the core of NeRF is a representation of a 3D scene as a volume whose views can be synthesized by querying a fully-connected neural network. 
However, NeRF has no control of individual semantic objects within a scene. Recently, GIRAFFE~\cite{niemeyer2021giraffe} proposed a generative model that represents objects in a compositional manner. While we propose a similar compositional approach, we tackle a different task of disentangling the foreground and background volumes of an existing scene and not generating them from scratch. This enables us to consider a wide variety of downstream manipulation tasks not possible in GIRAFFE. 
Lastly, recent work such as \cite{michel2021text2mesh,wang2021clip,sanghi2021clip} considered the ability to manipulate 3D scenes semantically using text. We demonstrate a similar capability, but one which transcends to individual objects or regions in our 3D scene, while adhering to the semantics of the background. Our task is therefore reminiscent of the 3D counterpart of 2D inpainting. 

Given a set of 2D training views and poses of a scene, as well as masks, specifying the foreground object, our method first trains a neural radiance field to reconstruct the background and its associated effects. We follow a similar procedure as NeRF~\cite{mildenhall2020representing} by training a fully connected neural network to predict a volume density and view-dependent emitted radiance for every 5D coordinate of position and viewing direction. Due to the prior induced through volumetric rendering, the resulting neural field captures the background volume that also includes objects appearing behind or occluding the foreground object, and captures associated effects such as illumination and reflections. The background and foreground can be rendered from both training and novel views. 
By training a neural radiance field to reconstruct the volume of the entire 3D scene and the volume of the background separately, the representation of the foreground can be computed in a compositional manner from the two volumes~\cite{drebin1988volume} as illustrated in \cref{fig:teaser}.

Having disentangled the foreground object from the rest of the 3D scene, we can now perform a range of downstream tasks. 
We note that certain AR settings are constrained to a specific type of modification of the original scene. For example, optical-see-through devices can only add light to the scene, meaning that the generation must be non-negative with respect to the input scene~\cite{luo2021stay}. In other cases, one may wish to keep the depth of the original scene intact~\cite{owens2014camouflaging,guo2022ganmouflage}, and only modify the textures or colors of objects. 
Our framework enables properties such as color, depth, and affine transformations of both the foreground object and background to be manipulated separately, and therefore can handle such manipulation tasks. To demonstrate this, we consider the task of object camouflage under constant depth and non-negative 3D inpainting. We also demonstrate the applicability of our method in transforming the foreground object separately from the background, while correctly accounting for illumination and reflection caused by the background. 

Lastly, we consider the 
ability to affect semantic manipulations to the foreground. 
To this end, we consider the recently proposed multi-modal embedding of CLIP~\cite{radford2021learning}. Using CLIP, we are able to manipulate the foreground object semantically using text. 
We note that while 2D counterparts may exist for each of the proposed manipulations, our disentangled volumetric manipulation offers 3D-consistent and semantic manipulation of foreground objects. Such manipulation successfully adheres to the global semantics of the scene while correctly capturing reflection and shading effects. 

To summarize, our work makes the following contributions: (i) We introduce a volumetric framework for disentangling foreground and background objects of an existing 3D scene, (ii) Given the disentangled 3D objects, we demonstrate semantically consistent manipulations of foreground and background objects for a wide range of downstream tasks of object camouflage, non-negative 3D object inpainting, 3D object translation, 3D object inpainting, and 3D text-based object manipulation.  We demonstrate these tasks in a range of experiments on a diverse set of natural scenes. 

\section{Related Work}
Our approach lies at the intersection of three lines of related work. We first consider related 3D representations suitable for 3D volumetric disentanglement. Then, we review related approaches that disentangle semantic proprieties in a 3D scene. Lastly, we examine other approaches for 3D manipulation tasks.

\subsection{Volumetric Representation}

The choice of object representation suitable for 3D disentanglement of volumetric objects is of particular importance.
Several representations can be used to learn 3D scenes such as point clouds \cite{Shu20193DPC,pointflow,Hui2020ProgressivePC,achlioptas2017latent_pc}, meshes \cite{MeshCNN,groueix2018,wang2018pixel2mesh,pan2019deep}, or voxels \cite{Riegler2017OctNet,xie2019pix2vox,NIPS2016_wu,brock2016}. However, work using these representations is typically restricted in topology or resolution.  

Another line of research implicitly represents the 3D volume to create a high-quality and scalable representation of 3D objects \cite{niemeyer2021giraffe,saito2019pifu,Park2019DeepSDFLC,Oechsle2019TextureFL,Niemeyer2019OccupancyF4}.  
Recently NeRF~\cite{mildenhall2020representing} proposed to apply an implicit model with volumetric differentiable rendering to synthesize high-fidelity and photo-realistic reconstructions of complex 3D scenes. 
Given a position and viewing direction, NeRF predicts a color and volume density using a fully connected neural network. A standard volume rendering pipeline can then use the output to render object views. 
However, NeRF cannot manipulate separate 3D volumetric entities within a scene. Our framework extends the representation of NeRF to allow for such separate control.

\subsection{Object Disentanglement}
Learning a disentangled representation of shapes, texture, and pose has been a subject of long-term interest in computer vision. 
We focus on volumetric disentanglement of semantic properties in 3D scenes. For a more comprehensive overview, please refer to \cite{Ahmed2018ASO}.
CLIP-NeRF~\cite{wang2021clip} disentangles the shape and appearance of NeRF and, subsequently, uses CLIP~\cite{radford2021learning} to manipulate these properties. However, it requires a synthetic object dataset for training and is limited to rigid objects, while our work can be used for complex scene manipulation in a zero-shot manner.
Other works disentangle pose~\cite{Wang2021NeRFNR,yen2020inerf}, illumination~\cite{nerv2021,boss2021nerd}, texture and shape~\cite{liu2021editing,Jang2021CodeNeRFDN}, but are limited to manipulating the entire volumetric scene. 

Perhaps most similar to our work is GIRAFFE~\cite{niemeyer2021giraffe}, which represents objects in a scene in a compositional manner. 
We differ from GIRAFFE in the following ways. (i) We disentangle an \textit{existing} scene into the foreground and background volumes, while GIRAFFE generates such volumes from scratch. This enables our method to tackle tasks including 3D object removal, 3D object camouflage, and 3D semantic manipulation of the foreground and background. 
(ii) GIRAFFE learns to predict a feature vector for every object given a position and viewing direction. Our method predicts the RGB color instead, and as such has separate control of the color of each object.
(iii) While GIRAFFE requires a large collection of images for training, our method requires a set of foreground masks and associated views from a single scene.
(iv) Lastly, our method enables zero-shot manipulations (does not require any 3D or 2D training data), can manipulate only the foreground or background separately, and adheres to the global semantics of the entire scene.

\subsection{3D Manipulation}

Our framework enables manipulation of localized regions in a scene. While 2D counterparts, such as approaches to 2D inpainting exist~\cite{guillemot2013image,yu2019free,efros1999texture,efros2001image}, they cannot generate 3D consistent manipulations. As we incorporate the prior of volumetric rendering, we naturally handle non-planar objects well as well as correlated effects such as lighting, reflections, and occlusions. Further, unlike our method, these methods require a large dataset of images with associated 2D masks. 

\textit{Global Scene Manipulation.}
One set of approaches considers editing the entire scene. \cite{canfes2022text} considers texture and shape manipulating of 3D meshes. CLIP-Forge \cite{Sanghi2021CLIPForgeTZ} generates objects matching a text prompt using CLIP embeddings.
Text2Mesh \cite{wang2021clip} holds geometry fixed and manipulates only the texture or style of an object~\cite{michel2021text2mesh}. 
DreamFields~\cite{jain2021zero} learn a neural radiance field representing 3D objects from scratch, in a generative manner. 
Unlike these works, our work is concerned with manipulating a local region in an existing scene.

\textit{Localized Object Manipulation.}
In the context of localized object manipulations, CodeNeRF~\cite{Jang2021CodeNeRFDN} and EditNeRF~\cite{liu2021editing} modify the shape and color code of objects using coarse 2D user scribbles, but require a curated dataset of objects under different colors and views, and are limited to synthetic objects. In contrast, our method enables manipulation of objects in complex scenes, semantically, according to a target text prompt.

\section{Method}

\begin{figure*}
    \centering
    \includegraphics[width=\textwidth]{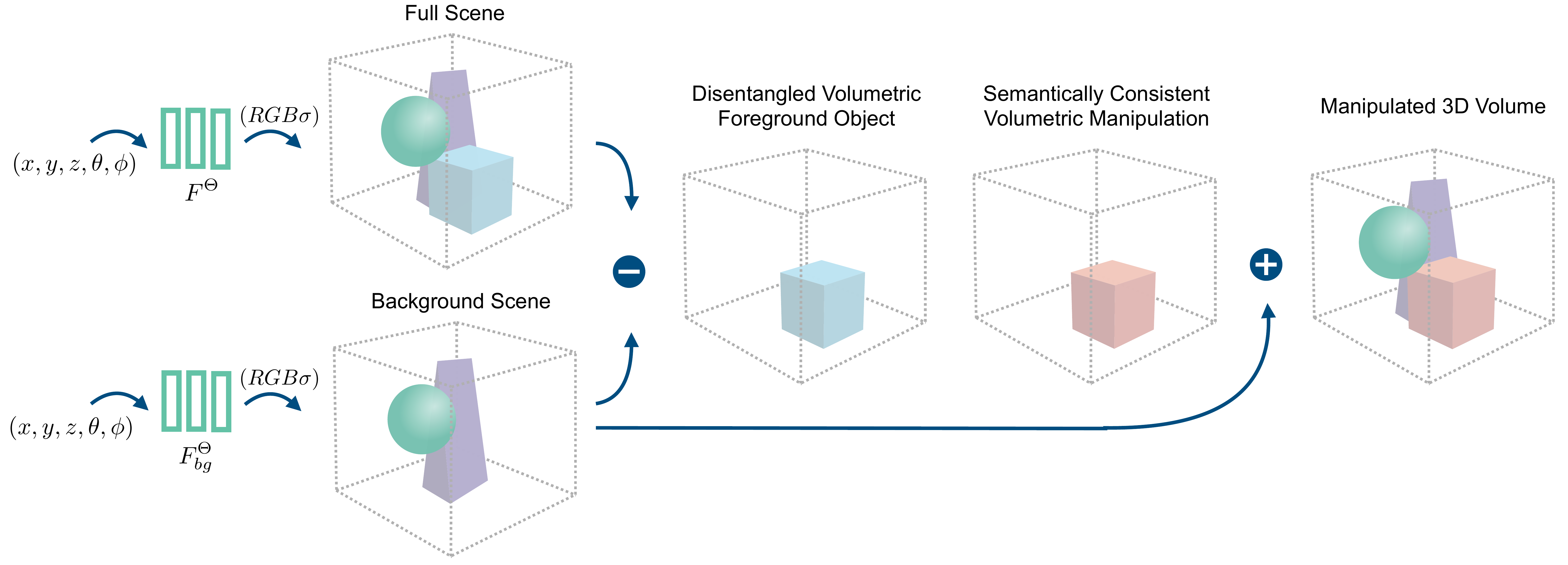}
    \caption{\textbf{Overview of our disentanglement framework.} First, we learn a volumetric representation of the background and full scene using neural radiance fields (\cref{sec:representation}). Second, by subtracting the full and the background volumes, we obtain a disentangled foreground volume. Third, we can then perform a wide range of manipulations on this volume, which adhere to the background volume. This is illustrated here by changing the color of the cube from blue to red. Finally, we can place the foreground object back into the original scene by adding it volumetrically to the background scene, obtaining a manipulated scene. 
    }
    \vspace{-0.3cm}
    \label{fig:disentanglement_overview}
\end{figure*}

Given a 3D scene, we wish to disentangle semantic objects from the rest of the scene. 
This enables the subsequent control of semantic objects as well as the background in a separate, but semantic manner. We focus on the setting of disentangling a single foreground object from the background, but our method easily extends to multiple objects. First, we describe the 3D volumetric representation used to disentangle objects and control objects separately (\cref{sec:representation}). The disentanglement of foreground and background volumes opens a wide range of downstream applications. We provide a framework that explores some of these applications by manipulating objects in a semantic manner (\cref{sec:manipulation}). 
An illustration of our framework is provided in \cref{fig:disentanglement_overview}. 

\subsection{Disentangled Object Representation}
\label{sec:representation}

The ability to disentangle the foreground object volumetrically from the background requires a volumetric representation that correctly handles multiple challenges:
(i). Foreground occluding objects, which may be covered by a foreground mask, should not be included in the foreground volume, (ii). Regions occluded by the foreground object should be visible in the background volume, (iii). Illumination and reflectance effects, affecting the foreground object in the full scene volume, should affect the now unoccluded regions of the background in a natural way. As shown in \cref{sec:results}, our framework correctly handles these challenges. To this end, we build upon the representation of neural radiance fields~\cite{mildenhall2020representing}.

\paragraph{Neural Radiance Fields.}
\label{sec:nerf}
A neural radiance field~\cite{mildenhall2020representing} is a continuous function $f$ whose input is a 3D position $p = (x, y, z) \in \mathbb{R}^3$ along with a viewing direction $d =(\theta, \phi) \in \mathbb{S}^2$, indicating a position along a camera ray. The output of $f$ is an RGB color $c \in \mathbb{R}^3$ and volume density $\alpha \in \mathbb{R}^{+}$.   
We follow \cite{mildenhall2020representing,tancik2020fourier} in first applying a frequency-based encoding $\gamma$ to correctly capture high frequency details:
\begin{align}
\gamma\left(p\right) = \left[\cos\left(2 \pi \mathbf Bp\right), \sin\left(2 \pi \mathbf Bp\right)\right]^\transpose
\label{eq:positional}
\end{align}
where $\mathbf{B} \in \mathbb{R}^{n \times 3}$ is a randomly drawn Gaussian matrix whose entries are drawn from $\mathcal{N}\left(0,\sigma^2\right)$, where $\sigma$ is an hyperparameter. $f$ is then paramaterized as an MLP $f_{\theta}$ whose input is $(\gamma(p), \gamma(d))$ and output is $c$ and $\sigma$. 

\paragraph{Object Representation.}

\begin{figure*}
    \centering
    \includegraphics[width=\textwidth]{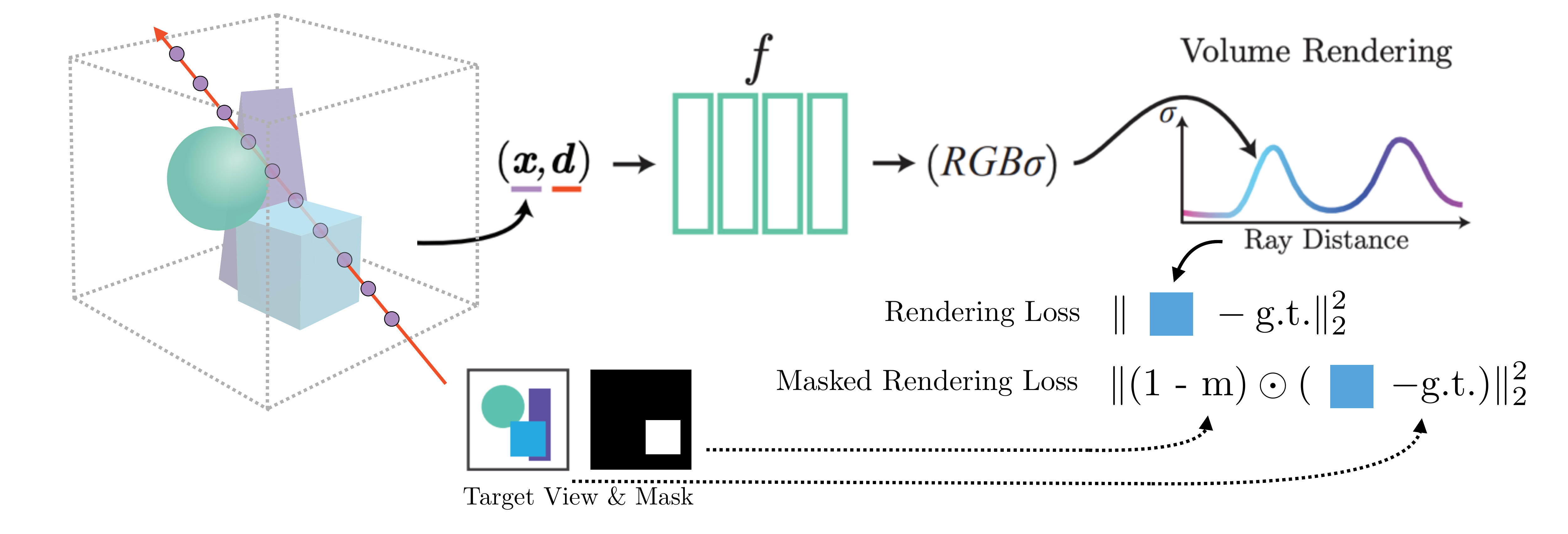}
    \caption{\textbf{Training losses for the background and full scene.} We train a neural radiance field for the \textit{full scene} with rendering loss on all ground truth pixels. To train the \textit{background scene}, we apply a masked rendering loss, where regions that are projected inside a 2D mask $(1 - m)$, are not penalized in the loss. The network learns to reconstruct this region based on correlated effects, such as how light from the surrounding affects the masked regions, and multi-view geometry, where the background might not be masked from another view. }
    \label{fig:loss_overview}
    \vspace{-0.2cm}
\end{figure*}

\label{sec:object}
Given camera extrinsics $\xi$, we assume a set $\{(c_r^i, \sigma_r^i) \}_{i=1}^N$ of color and volume density values predicted by $f_{\theta}$ for $N$ randomly chosen points along camera ray $r$. A rendering operator 
then maps these values to an RGB color $c_r$ as follows:
\begin{align}
    c_r = \sum_{i=1}^N w^i_r \cdot c^i_r \hspace{1cm} w_r^i = \prod_{j=1}^i (1 - \sigma^r_j) \cdot T^i_r \hspace{1cm} T^i_r = 1 - \exp(\sigma^i_r \cdot \delta^r_i)
    \label{eq:rendering} 
\end{align}
where $\sigma^i_r$ and $T^i_r$ are the alpha and transmittance values for point $i$ 
along ray $r$ and $\delta^r_i = t^{i+1} - t^i$ is the distance between adjacent samples. 

For training, we assume a set of posed views $\{x_i\}_{i=1}^M$ 
together with their associated foreground object masks $\{m_i\}_{i=1}^M$. We set  $\{\hat{x_i}\}_{i=1}^M$ to be the corresponding colors to $\{x_i\}_{i=1}^M$ as predicted 
by \cref{eq:rendering}. 
\cref{fig:loss_overview} gives an overview of the training. To train the background volume, we minimize the masked reconstruction loss between real and estimated views:
\begin{align}
    \mathcal{L}_{bg} = \sum_{i=1}^M || (1 - m_i) \odot (x_i - \hat{x}_i)||_2^2
\end{align}
A volume of the full scene is also trained with the following loss:
\begin{align}
    \mathcal{L}_{full} = \sum_{i=1}^M || x_i - \hat{x}_i ||_2^2
    \label{eq:full}
\end{align}
Let $w_{bg}^{i_r}$ and $c_{bg}^{i_r}$ be the value of $w^i_r$ and $c^i_r$ in \cref{eq:rendering} predicted for the background volume and similarly let $w_{full}^{i_r}$ and $c_{full}^{i_r}$ be the value of $w^i_r$ and $c^i_r$ in \cref{eq:rendering} predicted for the full volume. 
A natural representation of the foreground object can then be found using the principle of volume mixing~\cite{drebin1988volume}: 
\begin{align}
    c^{fg}_r = \sum_{i=1}^N w_{fg}^{i_r} \cdot c_{fg}^{i_r}  \hspace{1cm} w_{fg}^{i_r} = w_{full}^{i_r} - w_{bg}^{i_r} \hspace{1cm} c_{fg}^{i_r} = c_{full}^{i_r} - c_{bg}^{i_r} 
    \label{eq:foreground}
\end{align}
$c^{fg}_r$ is the foreground volume color at the pixel corresponding to ray $r$. \cref{eq:foreground} renders the color of the foreground object for all pixels across different views.

\paragraph{Object Controls.}
\label{sec:objectcontrols}

We note that camera parameters, as well as chosen poses, rays, and sampled points along the rays are chosen to be identical for both the full volume and the background volume, and hence also identical to the foreground volume. Given this canonical setting, the corresponding points along the rays for both the foreground and background can be easily found. 

Due to the above mentioned correspondence, one can independently modify ${w_{fg}}^{i_r}$ and ${c_{fg}}^{i_r}$ to get
${w'_{fg}}^{i_r}$ and ${c'_{fg}}^{i_r}$ for the foreground volume as well as  ${w_{bg}}^{i_r}$ and ${c_{bg}}^{i_r}$ 
to get ${w'_{bg}}^{i_r}$ and ${c'_{bg}}^{i_r}$ for the background volume. In order to recombine the modified background with the modified foreground, we note that every 3D point along the ray should only be colored, either according to the background volume or according to the foreground volumes, but not by both, as they are disentangled. 
We can then recombine the modified foreground and background: 
\begin{align}
    c^{c}_r = \sum_{i=1}^N {w'_{bg}}^{i_r} \cdot {c'_{bg}}^{i_r} + {w'_{fg}}^{i_r} \cdot {c'_{fg}}^{i_r}  
    \label{eq:composition}
\end{align}
$c^{c}_r$ is the recombined color of the pixel corresponding to ray $r$. In our experiments, we only modify the foreground and so ${w'_{bg}}^{i_r} = {w_{bg}}^{i_r}$, ${c'_{bg}}^{i_r} = {c_{bg}}^{i_r}$.

\subsection{Object Manipulation}
\label{sec:manipulation}

Given the ability to control the foreground and background volumes separately, we now propose a set of downstream manipulation tasks that emerge from our disentangled representation. As noted in \cref{sec:objectcontrols}, we can now control the weights, colors as well as translation parameters separately for the foreground and background volumes and so introduce a set of manipulation tasks that use the controls. We note that the task of \textit{Object Removal} is equivalent to displaying the background volume in isolation.

\paragraph{Object Transformation.}
Due to the alignment of camera parameters, as well as chosen poses, rays, and sampled points along the rays, one can apply a transformation on the background and foreground volumes separately, before recombining the volumes together. For either the foreground or the background, and for a given transformation $T$ (for example, a rotation or translation), we simply evaluate the color and weight of point $p$ using $f_{\theta}$ at position $T^{-1}(p)$ and then recombine the volumes together using \cref{eq:composition}. 

\paragraph{Object Camouflage.} 
Here we wish to change the texture of the foreground 3D object such that it is difficult to detect from its background~\cite{owens2014camouflaging,guo2022ganmouflage}. Such an ability can be useful in the context of diminished reality~\cite{mori2017survey}. To do so, we fix the depth of the foreground object while manipulating its texture. As the depth of the foreground is derived from ${w_{fg}}^{i_r}$, we fix ${w'_{fg}}^{i_r} = {w_{fg}}^{i_r}$ and only optimize ${c'_{fg}}^{i_r}$. 
We follow
\cref{eq:composition}, in compositing the foreground and background volumes. Let the resulting output for each view $i$ be $\hat{x}^{c}_i$, and let $\hat{x}^{bg}_i$ be the corresponding output for the background volume. We optimize a neural radiance field for foreground colors ${c'_{fg}}^{i_r}$ with the following loss:
\begin{align}
    \mathcal{L}_{camouflage} = \sum_{i=1}^M || \hat{x}^{c}_i - \hat{x}^{bg}_i ||_2^2
\end{align}
As the depth is fixed, only the foreground object colors are changed so as to match the background volume as closely as possible. 

\paragraph{Non-negative 3D Inpainting.} Next, we consider the setting of non-negative image generation~\cite{luo2021stay}. In this setting, we are interested in performing non-negative changes to views of the full scene so as to most closely resemble the background volume. This constraint is imposed in optical-see-through devices that can only add light onto an image. In this case, we learn a residual volume to render views $\hat{x}^{residual}_i$ as in \cref{eq:rendering} with the following loss:
\begin{align}
    \mathcal{L}_{non-negative} = \sum_{i=1}^M || \hat{x}^{full}_i + \hat{x}^{residual}_i - \hat{x}^{bg}_i ||_2^2
\end{align}
where $\hat{x}^{full}_i$ are rendered views of the full scene as in \cref{eq:full}. That is, we learn a residual volume whose views are $\hat{x}^{residual}_i$, such that when added to the full volume views, most closely resemble the views of the background. 

\paragraph{Semantic Manipulation.} Next, we consider a mechanism for the semantic manipulation of the foreground, while adhering to the global semantics of the entire scene. 
To this end, we consider the recently proposed model of CLIP~\cite{radford2021learning}, a multi-modal embedding method that can be used to find the perceptual similarity between images and texts.
One can use CLIP to embed an image $I$ and text prompt $t$, and to subsequently compare the cosine similarity of the embeddings. Let this operation be $sim(I, t)$, where a value of $1$ indicates perceptually similar of a text and image. 
We note that one can also use CLIP to compare the perceptual similarity of two images $I_1$ and $I_2$, denoted $sim(I_1, I_2)$. 
Let $\hat{x}^{c}_i$ be the result of applying \cref{eq:composition}, while fixing the background colors and weights as well as the foreground weights. That is, we only optimize the foreground colors ${c'_{fg}}^{i_r}$. For a user-specified target text $t$, we consider the following objective:
\begin{align}
   \mathcal{L}_{semantic} =&  \sum_{i=1}^M 1 - sim \left( \hat{x}^{c}_i \odot m_i + 
   \hat{x}^{bg}_i \odot (1 - m_i), t \right)  \label{eq:semantic_1} \\
   &+ 1 - sim \left( \hat{x}^{c}_i \odot m_i + 
   \hat{x}^{bg}_i \odot (1 - m_i), \hat{x}^{bg}_i \odot (1 - m_i) \right)  \label{eq:semantic_2} \\ 
   & + || \hat{x}^{c}_i \odot (1 - m_i), \hat{x}^{bg}_i \odot (1 - m_i)||_2^2 \label{eq:semantic_3}
\end{align}
We note that while only the colors of the foreground volume can be manipulated, we enforce that such changes only occur within the localized masked region of the foreground, and so take the background from the fixed background volume.
To do so instead of applying clip similarity directly with $\hat{x}^{c}_i$, we apply it with $\hat{x}^{c}_i \odot m_i + 
   \hat{x}^{bg}_i \odot (1 - m_i)$.
Therefore, CLIP's similarity can only be improved by making local changes that occur within the masked region of the foreground object, but can 'see' the background as well as the foreground for context. We enforce the generated volume views are similar to both the target text (\cref{eq:semantic_1}) and the background (\cref{eq:semantic_2}). 
To further enforce that no changes are made to the background, we constrain the background of the combined volume views to match those of the background using \cref{eq:semantic_3}. 

\subsection{Training and Implementation Details}

For training, we consider the natural non-synthetic scenes given in \cite{mildenhall2020representing}, together with their associated pose information. An off-the-shelf segmentation or manual annotation is used to extract masks. We note that masks need not be exact, and may capture more then the desired object (see main paper for details). 
Our rendering resolution for training the background and full scenes is $504 \times 378$. For the manipulation tasks, the same resolution is used for \textit{3D inpainting}, \textit{object camouflage}, \textit{transformation} and \textit{non-negative inpainting} tasks. For the \textit{semantic manipulation} task, our rendering resolution is $252 \times 189$. For the CLIP~\cite{radford2021learning} input, for a given view, we sample a $128 \times 128$ grid of points from the $252 \times 189$ output, and then upsample it to $224 \times 224$, which is the required input resolution of CLIP. We normalize the images and apply a text and image embedding as in  CLIP~\cite{radford2021learning}. We follow NeRF~\cite{mildenhall2020representing}, in optimizing both a ``coarse" and ``fine" networks for a neural radiance field, and follow the same sampling strategy of points along the ray. 
All neural fields are parametrized using an MLP with ReLU activation of the same architecture of 
\cite{mildenhall2020representing}. We use  an Adam optimizer ($\beta_1 = 0.9$, $\beta_2 = 0.999$) with a learning rate that begins with $5 \times 10^{-4}$ and decays exponentially to $5 \times 10^{-5}$.

\section{Experiments}
\label{sec:results}

We divide the experimental section into two parts. First, we show that we can successfully disentangle the foreground and background volumes from the rest of the scene. Second, we demonstrate some of the many manipulation tasks this disentanglement enables, as described in \cref{sec:manipulation}. Corresponding 3D scenes from multiple views are provided in the project webpage. 
As far as we can ascertain, no other framework enables all applications we consider at once, in a simple and intuitive manner. However, fer specific applications we consider corresponding baselines.  

\begin{table}[t]
\centering
\begin{tabular}{l@{~~~}c@{~~}c@{~~}c@{~~}c@{~~}c@{~~}c}
\toprule
& \multicolumn{3}{c}{Object Disentanglement} &  \multicolumn{3}{c}{Object Manipulation} \\ 
\cmidrule(l){2-4}\cmidrule(l){5-7}  
   & Ours                     & DeepFill-v2~\cite{yu2019free}             & EdgeConnect~\cite{nazeri2019edgeconnect}                           & Ours                     & GLIDE~\cite{nichol2021glide}                    & Blended~\cite{avrahami2021blended}                  \\
   \midrule
Q1 & \textbf{3.86} & 2.44 & 2.37 &  \textbf{3.85} & 1.10 & 1.26 \\
Q2 & \textbf{3.84} & 1.52 & 1.86 &  \textbf{3.78} & 1.20 & 1.26  \\
\bottomrule
\end{tabular}

\caption{A user study performed for the tasks of Object Disentanglement and 3D Object Manipulation. A mean opinion score (1-5) is shown where users were asked: (Q1) ``How well was the desired task performed?'' (object removed or semantically manipulated) and (Q2) ``How realistic is the resulting scene?'' }
\label{tab:user_study}
\end{table}

\subsection{Object Disentanglement}
\label{sec:object_disent}

\begin{figure*}
     \centering
     \begin{subfigure}[b]{0.15\textwidth}
         \centering
         \includegraphics[width=\linewidth]{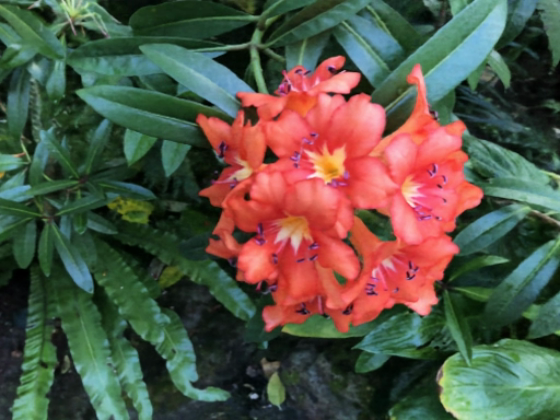}
     \end{subfigure}
     \begin{subfigure}[b]{0.15\textwidth}
         \centering
         \includegraphics[width=\linewidth]{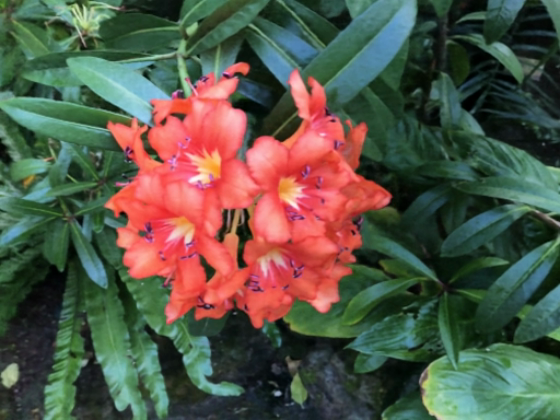}
     \end{subfigure}
     \begin{subfigure}[b]{0.15\textwidth}
         \centering
         \includegraphics[width=\linewidth]{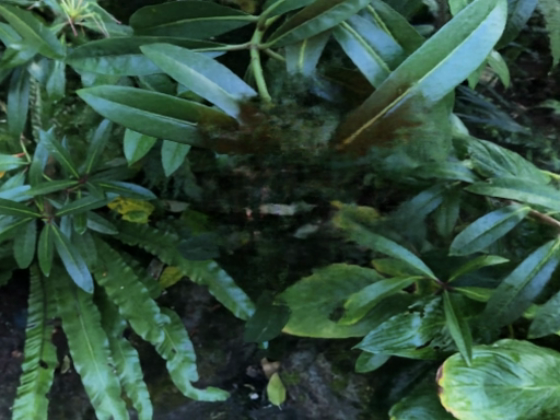}
     \end{subfigure} 
     \begin{subfigure}[b]{0.15\textwidth}
         \centering
         \includegraphics[width=\linewidth]{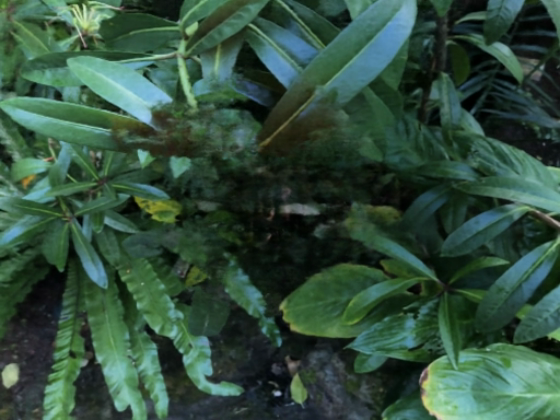}
     \end{subfigure}
     \begin{subfigure}[b]{0.15\textwidth}
         \centering
         \includegraphics[width=\linewidth]{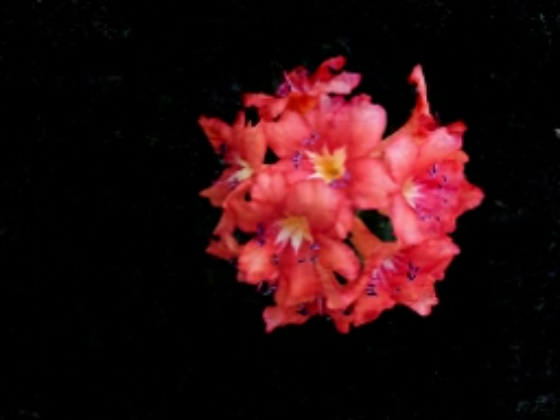}
     \end{subfigure}
     \begin{subfigure}[b]{0.15\textwidth}
         \centering
         \includegraphics[width=\linewidth]{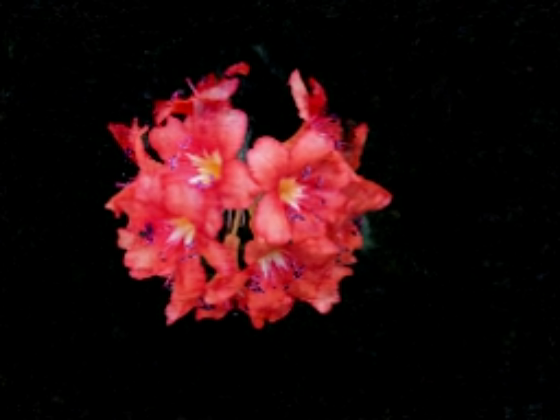}
     \end{subfigure}\\
     
     \begin{subfigure}[b]{0.15\textwidth}
         \centering
         \includegraphics[width=\linewidth]{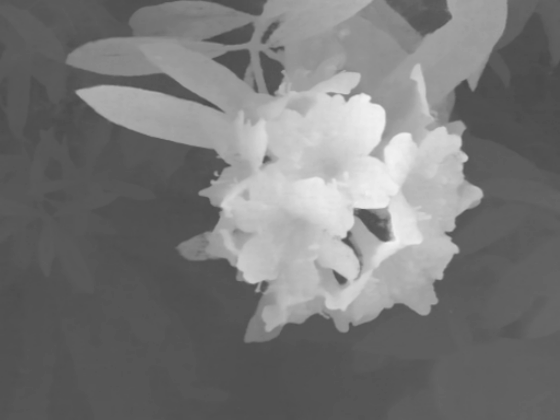}
     \end{subfigure}
     \begin{subfigure}[b]{0.15\textwidth}
         \centering
         \includegraphics[width=\linewidth]{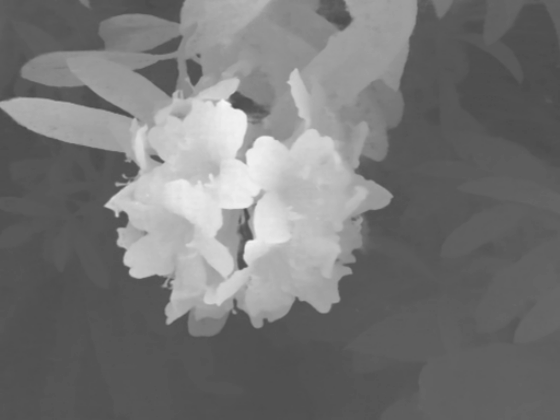}
     \end{subfigure}
     \begin{subfigure}[b]{0.15\textwidth}
         \centering
         \includegraphics[width=\linewidth]{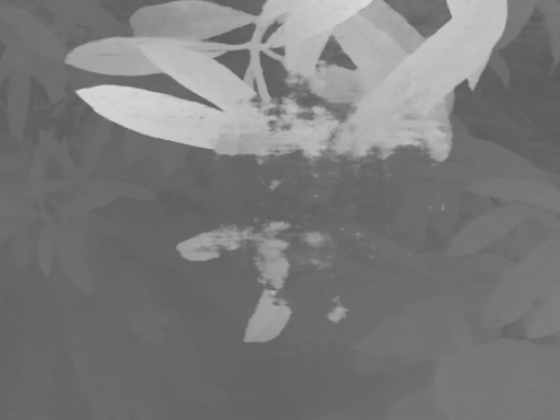}
     \end{subfigure} 
     \begin{subfigure}[b]{0.15\textwidth}
         \centering
         \includegraphics[width=\linewidth]{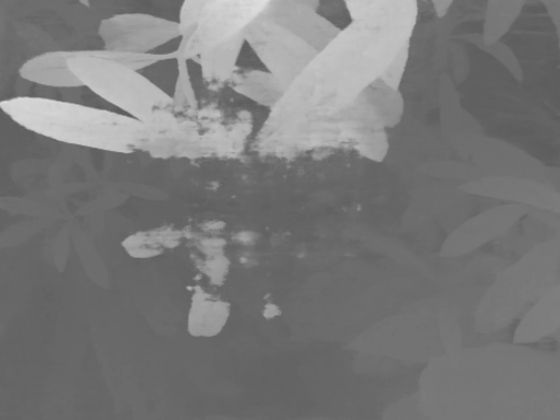}
     \end{subfigure}
     \begin{subfigure}[b]{0.15\textwidth}
         \centering
         \includegraphics[width=\linewidth]{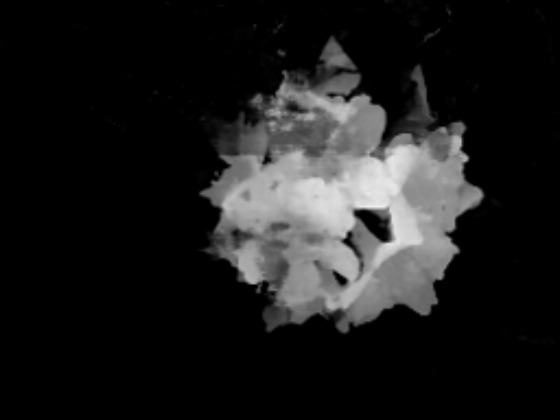}
     \end{subfigure}
     \begin{subfigure}[b]{0.15\textwidth}
         \centering
         \includegraphics[width=\linewidth]{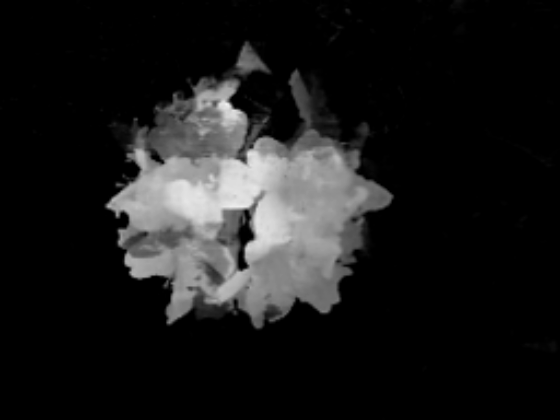}
     \end{subfigure}\\

     \begin{subfigure}[b]{0.15\textwidth}
         \centering
         \includegraphics[width=\linewidth]{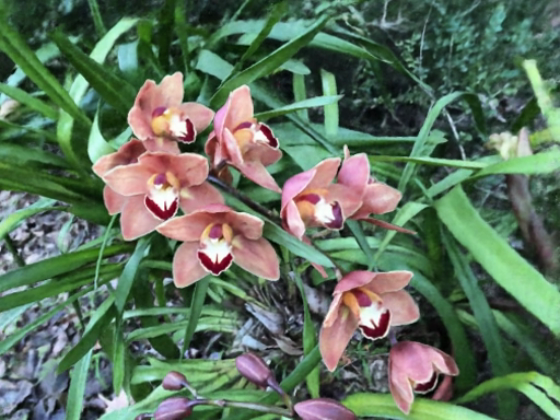}
     \end{subfigure}
     \begin{subfigure}[b]{0.15\textwidth}
         \centering
         \includegraphics[width=\linewidth]{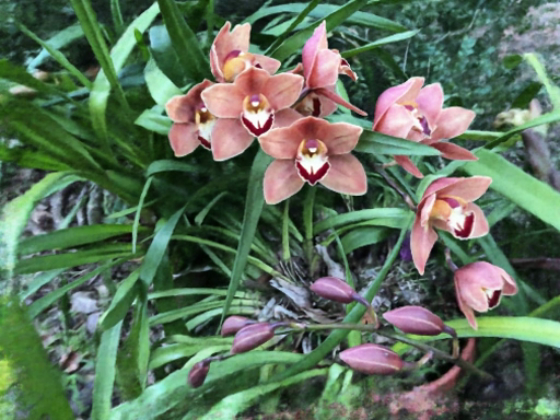}
     \end{subfigure}
     \begin{subfigure}[b]{0.15\textwidth}
         \centering
         \includegraphics[width=\linewidth]{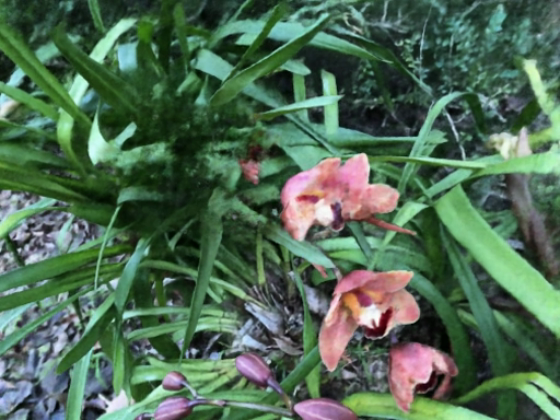}
     \end{subfigure} 
     \begin{subfigure}[b]{0.15\textwidth}
         \centering
         \includegraphics[width=\linewidth]{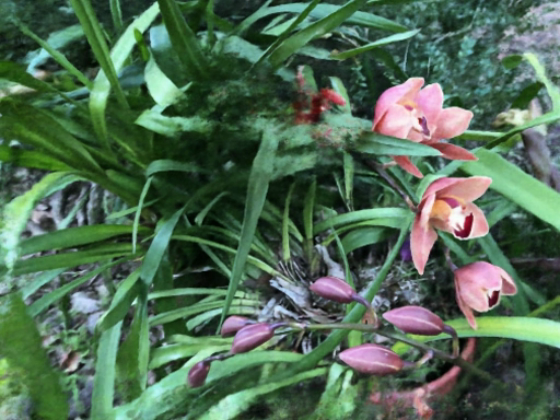}
     \end{subfigure}
     \begin{subfigure}[b]{0.15\textwidth}
         \centering
         \includegraphics[width=\linewidth]{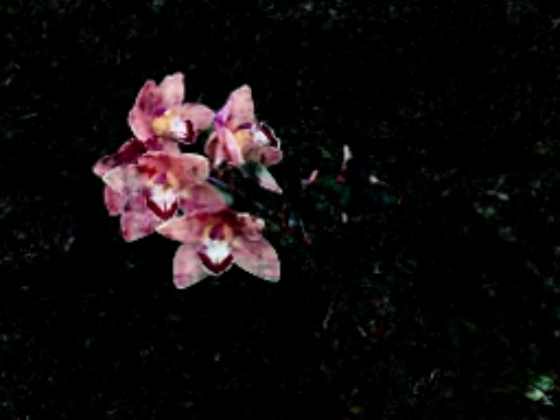}
     \end{subfigure}
     \begin{subfigure}[b]{0.15\textwidth}
         \centering
         \includegraphics[width=\linewidth]{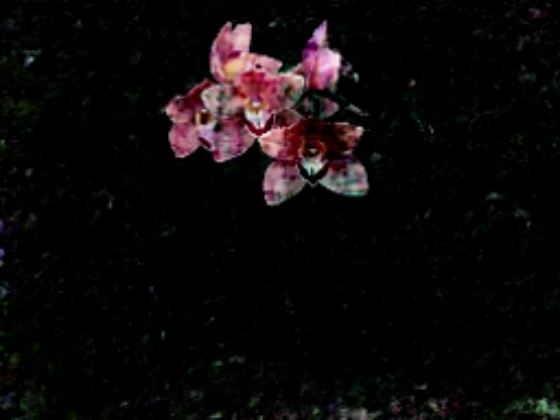}
     \end{subfigure}\\
     
     \begin{subfigure}[b]{0.15\textwidth}
         \centering
         \includegraphics[width=\linewidth]{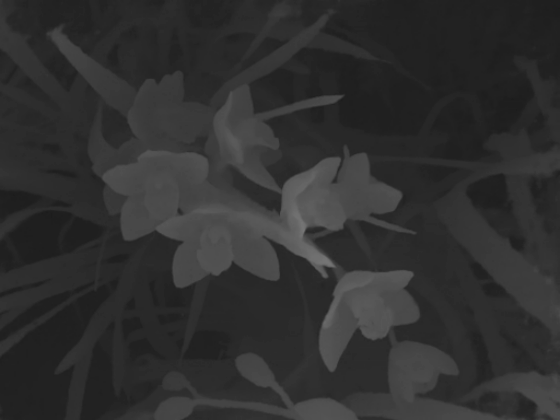}
     \end{subfigure}
     \begin{subfigure}[b]{0.15\textwidth}
         \centering
         \includegraphics[width=\linewidth]{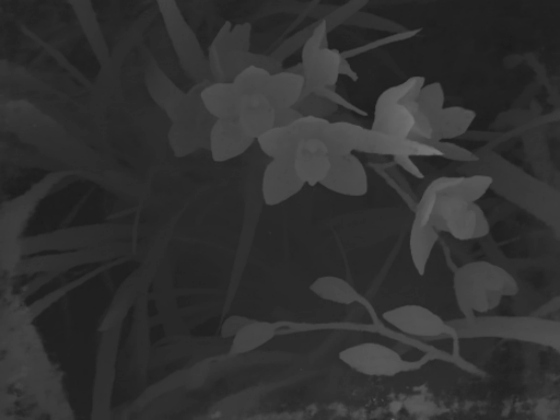}
     \end{subfigure}
     \begin{subfigure}[b]{0.15\textwidth}
         \centering
         \includegraphics[width=\linewidth]{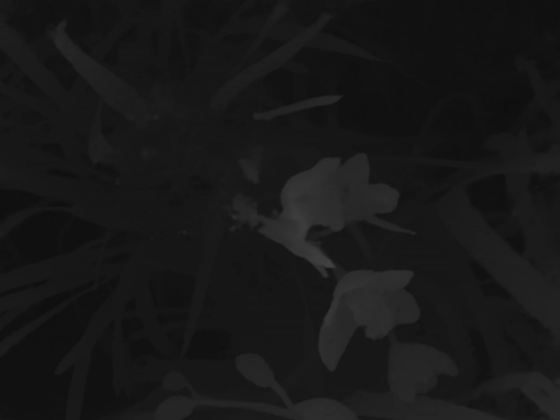}
     \end{subfigure} 
     \begin{subfigure}[b]{0.15\textwidth}
         \centering
         \includegraphics[width=\linewidth]{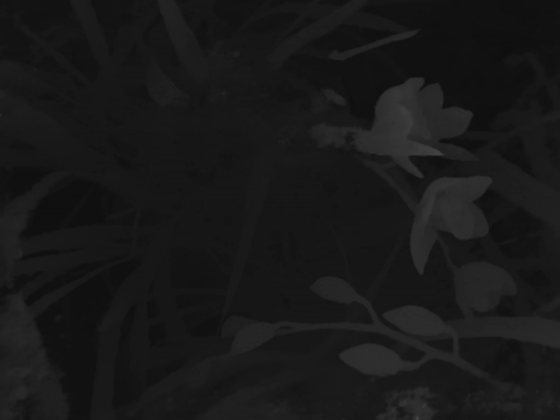}
     \end{subfigure}
     \begin{subfigure}[b]{0.15\textwidth}
         \centering
         \includegraphics[width=\linewidth]{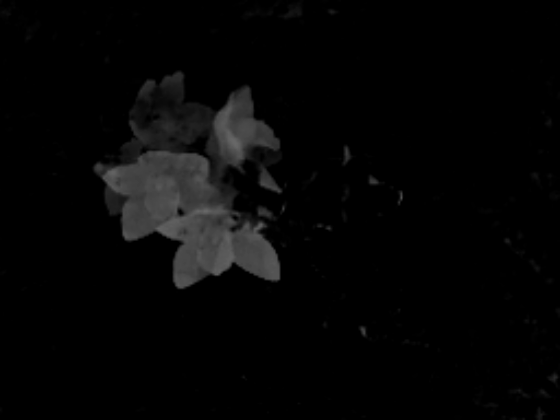}
     \end{subfigure}
     \begin{subfigure}[b]{0.15\textwidth}
         \centering
         \includegraphics[width=\linewidth]{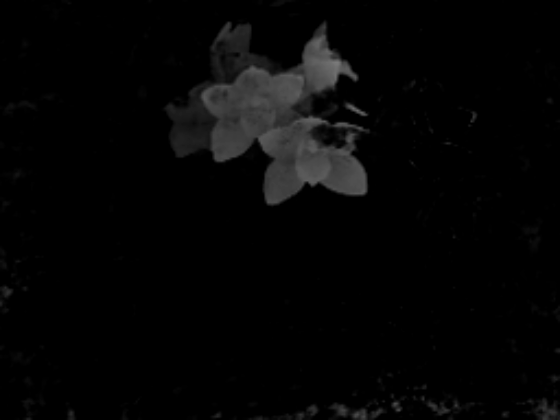}
     \end{subfigure}\\
     \hspace{0.5cm} Full Scene \hspace{1.5cm} Background Scene \hspace{1cm} Foreground Object
     \vspace{-0.1cm}
        \caption{\textbf{Two rendered views of the full scene, background and foreground.}
        As foreground is obtained by subtracting the background from the full scene volumetrically (\cref{sec:representation}), we also obtain the disparity of the foreground. }
        \label{fig:disentanglement_depth_maps}
        \vspace{0.1cm}

    \centering
     \begin{subfigure}[b]{0.182\textwidth}
         \centering
         \includegraphics[width=\linewidth]{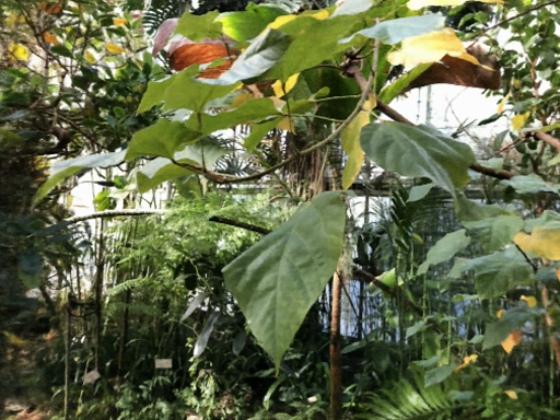}
     \end{subfigure}
     \begin{subfigure}[b]{0.182\textwidth}
         \centering
         \includegraphics[width=\linewidth]{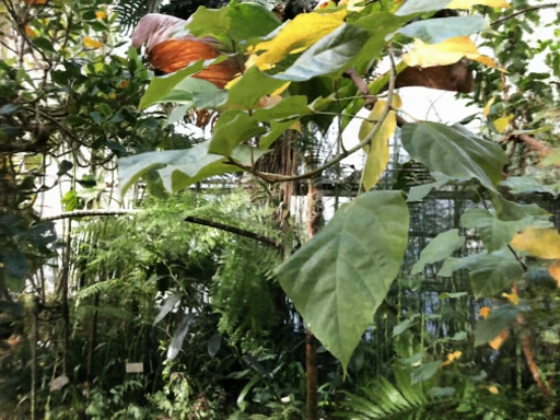}
     \end{subfigure}
     \begin{subfigure}[b]{0.182\textwidth}
         \centering
         \includegraphics[width=\linewidth]{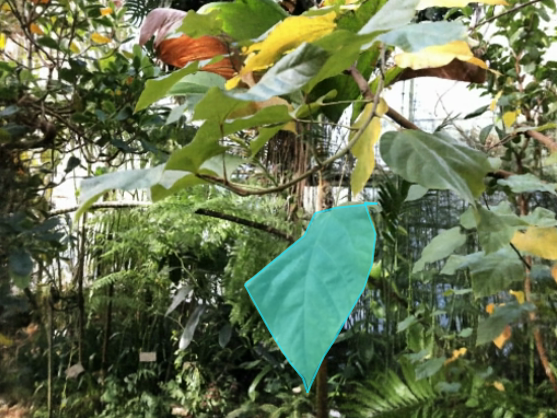}
     \end{subfigure} 
     \begin{subfigure}[b]{0.182\textwidth}
         \centering
         \includegraphics[width=\linewidth]{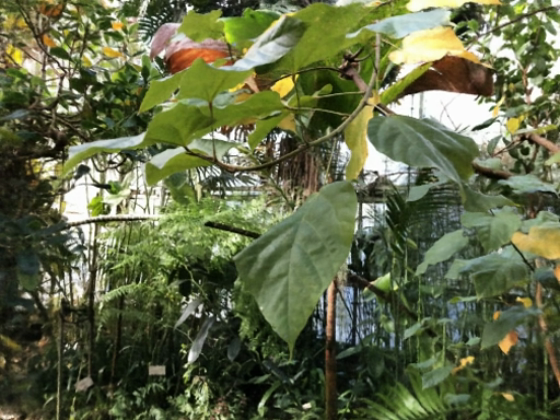}
     \end{subfigure}
     \begin{subfigure}[b]{0.182\textwidth}
         \centering
         \includegraphics[width=\linewidth]{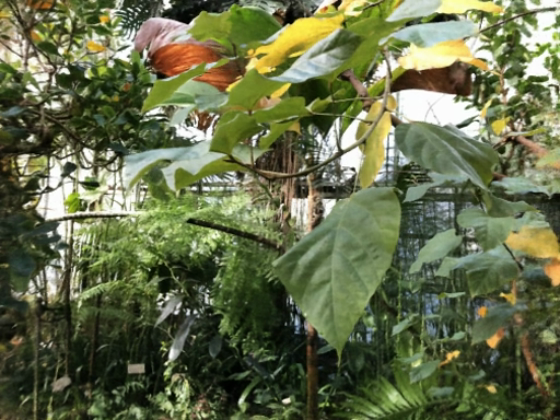}
     \end{subfigure}\\
     
     \begin{subfigure}[b]{0.182\textwidth}
         \centering
         \includegraphics[width=\linewidth]{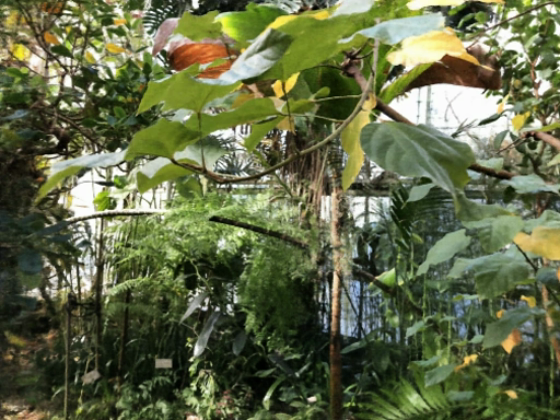}
     \end{subfigure}
     \begin{subfigure}[b]{0.182\textwidth}
         \centering
         \includegraphics[width=\linewidth]{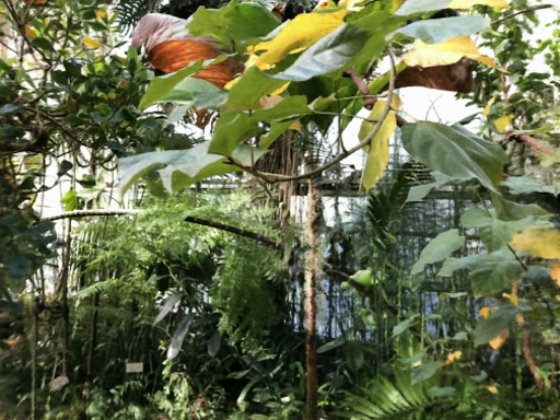}
     \end{subfigure} 
     \begin{subfigure}[b]{0.182\textwidth}
         \centering
         \includegraphics[width=\linewidth]{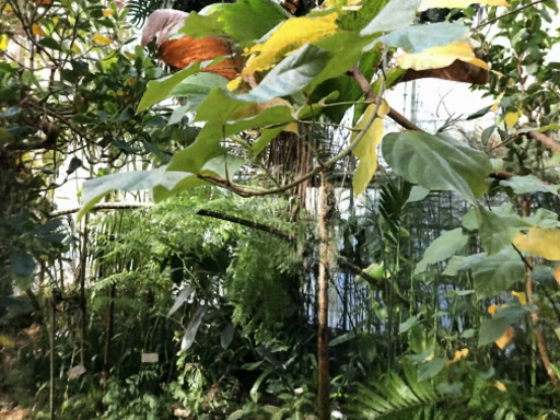}
     \end{subfigure}
     \begin{subfigure}[b]{0.182\textwidth}
         \centering
         \includegraphics[width=\linewidth]{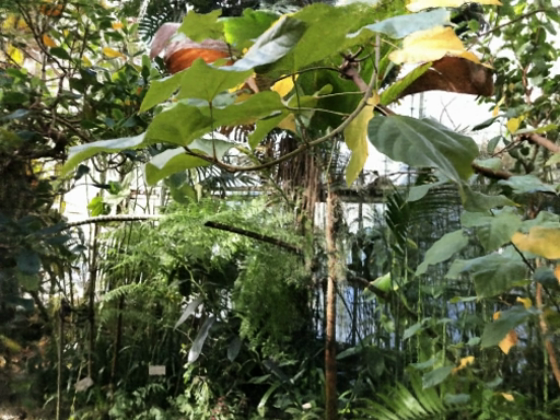}
     \end{subfigure}
     \begin{subfigure}[b]{0.182\textwidth}
         \centering
         \includegraphics[width=\linewidth]{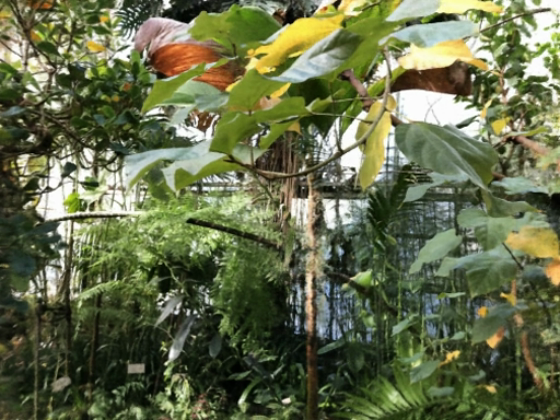}
     \end{subfigure}\\

     \begin{subfigure}[b]{0.182\textwidth}
         \centering
         \includegraphics[width=\linewidth]{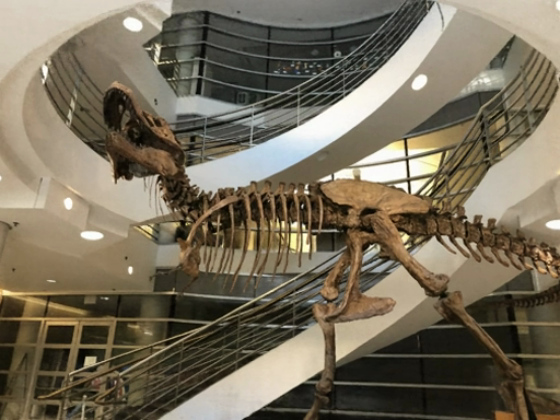}
     \end{subfigure}
     \begin{subfigure}[b]{0.182\textwidth}
         \centering
         \includegraphics[width=\linewidth]{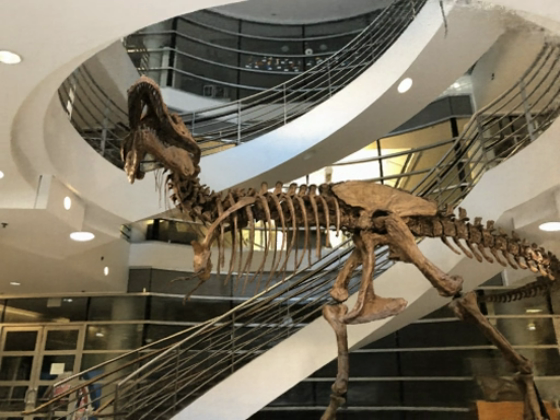}
     \end{subfigure}
     \begin{subfigure}[b]{0.182\textwidth}
         \centering
         \includegraphics[width=\linewidth]{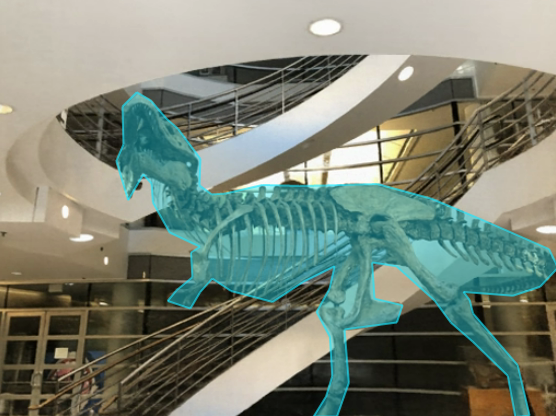}
     \end{subfigure} 
     \begin{subfigure}[b]{0.182\textwidth}
         \centering
         \includegraphics[width=\linewidth]{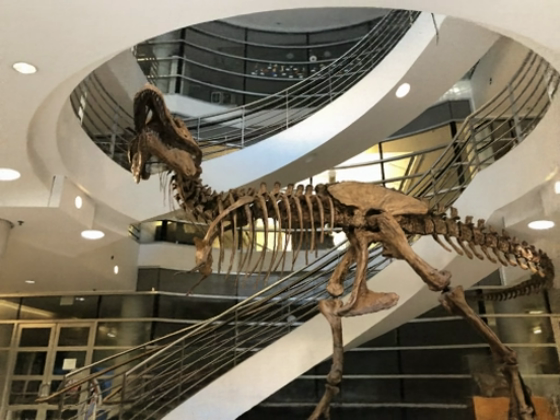}
     \end{subfigure}
     \begin{subfigure}[b]{0.182\textwidth}
         \centering
         \includegraphics[width=\linewidth]{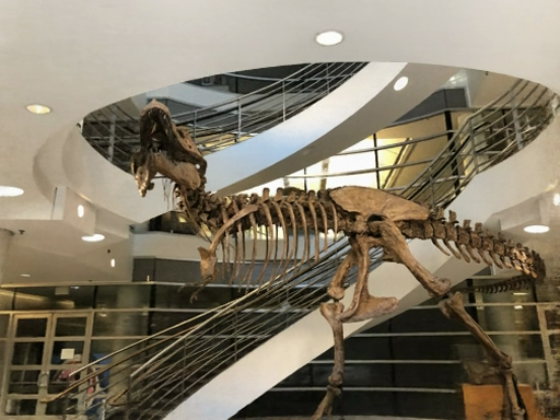}
     \end{subfigure}\\
     
     \begin{subfigure}[b]{0.182\textwidth}
         \centering
         \includegraphics[width=\linewidth]{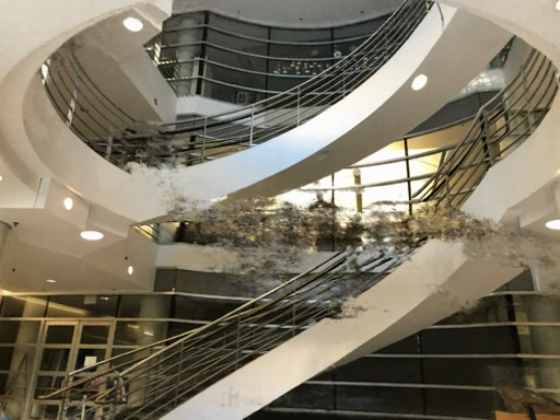}
     \end{subfigure}
     \begin{subfigure}[b]{0.182\textwidth}
         \centering
         \includegraphics[width=\linewidth]{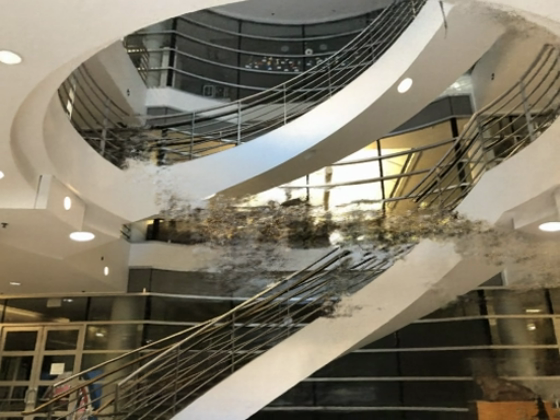}
     \end{subfigure} 
     \begin{subfigure}[b]{0.182\textwidth}
         \centering
         \includegraphics[width=\linewidth]{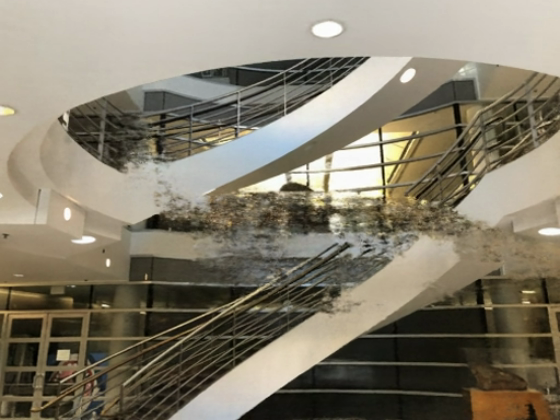}
     \end{subfigure}
     \begin{subfigure}[b]{0.182\textwidth}
         \centering
         \includegraphics[width=\linewidth]{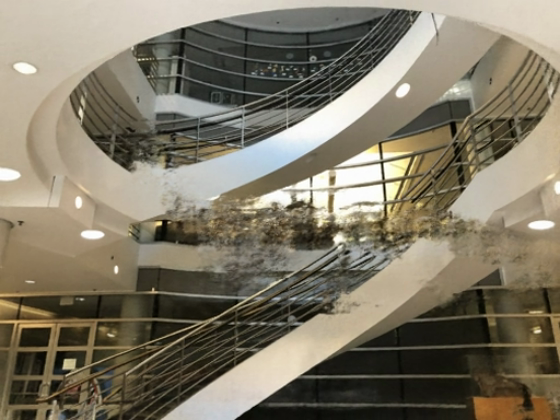}
     \end{subfigure}
     \begin{subfigure}[b]{0.182\textwidth}
         \centering
         \includegraphics[width=\linewidth]{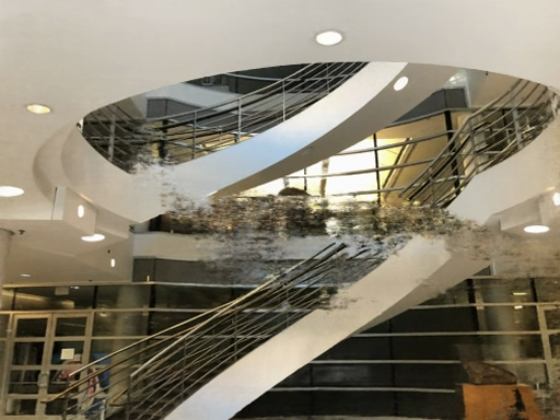}
     \end{subfigure}\\

     \begin{subfigure}[b]{0.182\textwidth}
         \centering
         \includegraphics[width=\linewidth]{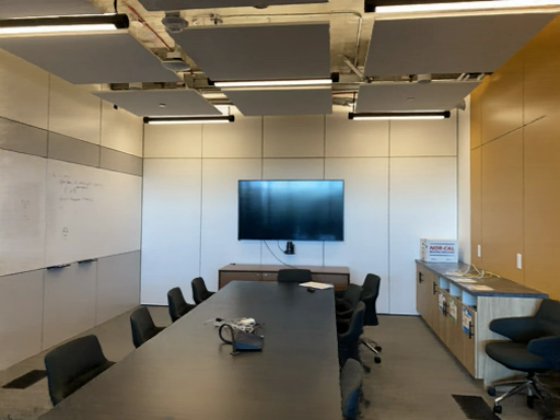}
     \end{subfigure}
     \begin{subfigure}[b]{0.182\textwidth}
         \centering
         \includegraphics[width=\linewidth]{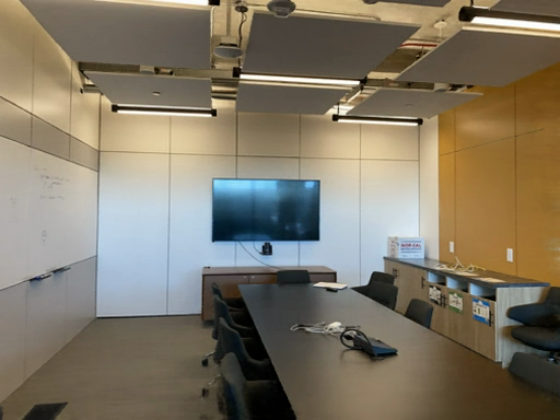}
     \end{subfigure}
     \begin{subfigure}[b]{0.182\textwidth}
         \centering
         \includegraphics[width=\linewidth]{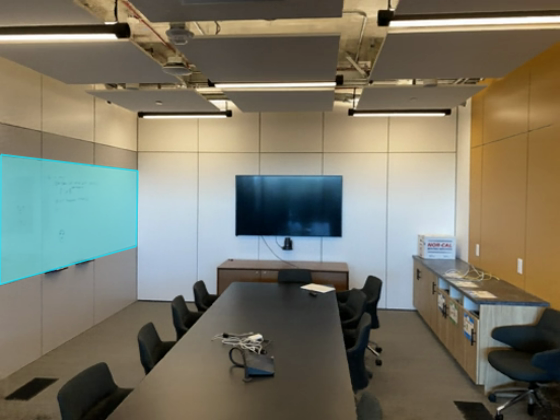}
     \end{subfigure} 
     \begin{subfigure}[b]{0.182\textwidth}
         \centering
         \includegraphics[width=\linewidth]{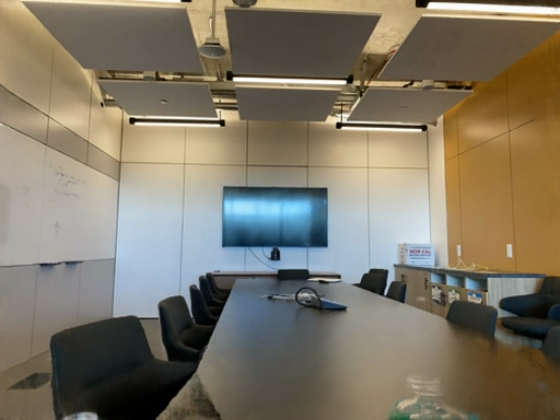}
     \end{subfigure}
     \begin{subfigure}[b]{0.182\textwidth}
         \centering
         \includegraphics[width=\linewidth]{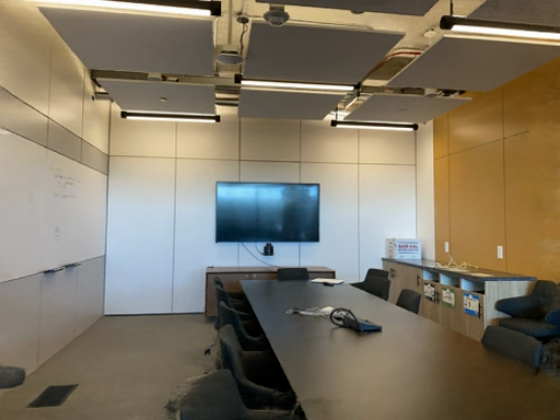}
     \end{subfigure}\\
     
     \begin{subfigure}[b]{0.182\textwidth}
         \centering
         \includegraphics[width=\linewidth]{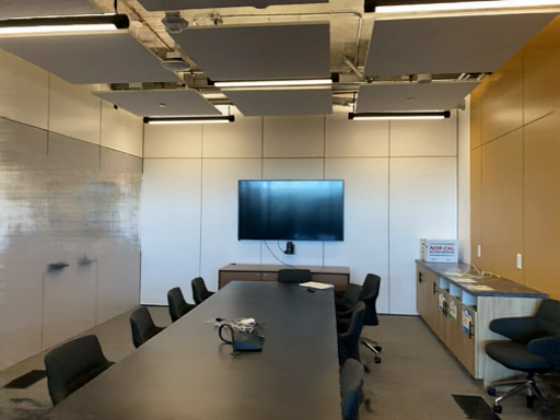}
     \end{subfigure}
     \begin{subfigure}[b]{0.182\textwidth}
         \centering
         \includegraphics[width=\linewidth]{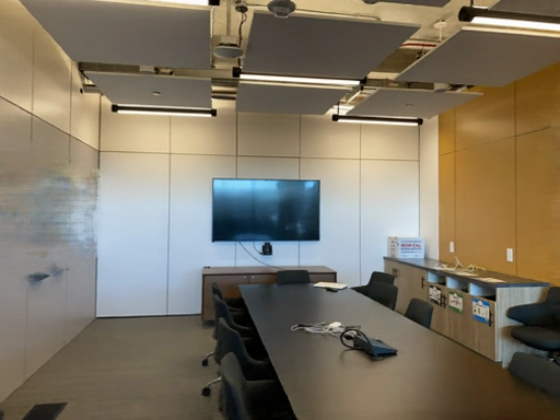}
     \end{subfigure} 
     \begin{subfigure}[b]{0.182\textwidth}
         \centering
         \includegraphics[width=\linewidth]{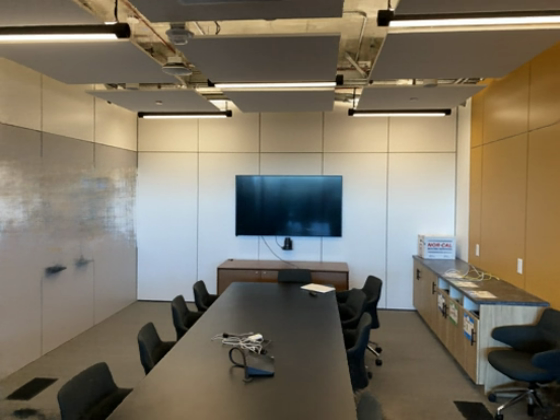}
     \end{subfigure}
     \begin{subfigure}[b]{0.182\textwidth}
         \centering
         \includegraphics[width=\linewidth]{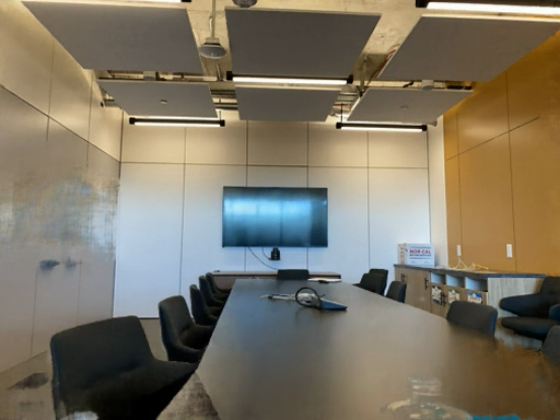}
     \end{subfigure}
     \begin{subfigure}[b]{0.182\textwidth}
         \centering
         \includegraphics[width=\linewidth]{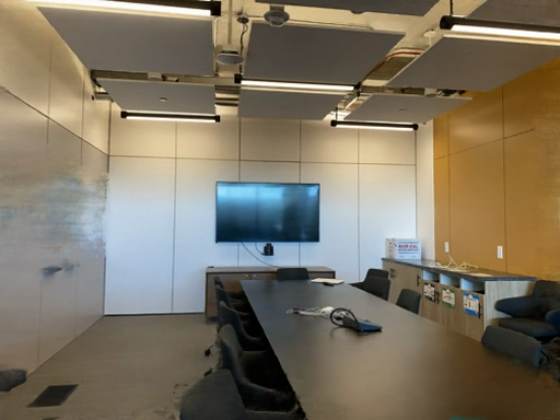}
     \end{subfigure}\\
        \caption{\textbf{Semantic consistency across views.}  Uniformly sampled renderings of the full and the background volumes for three different scenes. The removed object is visually enhanced in the center column by a 2D mask. } 
        \label{fig:disentanglement_view_consistency}
\end{figure*}

\cref{fig:disentanglement_depth_maps} shows views from different scenes where we separate the full scene,  background, and foreground in a volumetrically and semantically consistent manner. 
As can be seen, the disentangled objects are consistent across views.  \cref{fig:disentanglement_view_consistency} shows how the removal of a leaf, a T-rex, and a whiteboard is consistent across multiple views. The background neural radiance field is made plausible predictions of the background scene via multi-view geometry and based on the correlated effects from the scene. \textit{E.g.} the background behind the leaf or the legs of the T-rex might be occluded by the 2D mask from one view, but visible from another. However, the background behind the whiteboard is occluded from every angle. Nevertheless, the background neural radiance field makes a plausible prediction of the background based on the correlated effects from the surrounding scene. Further, our model can handle the disentanglement of non-planar objects, such as the T-rex, well.

As far as we can ascertain, the closest 2D task to object disentanglement is that of object inpainting. We consider two prominent baselines of DeepFill-v2~\cite{yu2019free} and EdgeConnect~\cite{nazeri2019edgeconnect} for this task and compare our method on the scenes of leaves and whiteboard removal as in Fig.~\ref{fig:disentanglement_view_consistency}.  
We train the baseline on the same training images and their associated masks. In order to compare our method on the same novel views, we train a NeRF~\cite{mildenhall2020representing} on the resulting outputs, resulting in a scene with the same novel views as ours. A visual comparison is given in the project webpage. Unlike our method, the results have 3D inconsistencies, artifacts and flickering between views. We note that as no ground truth views exist, standard metrics such as PNSR/SSIM are not applicable. 
We therefore conduct a user study and ask users to rate from a scale of $1-5$: (Q1) ``How well was the object removed?'' and (Q2) ``How realistic is the resulting scene?'' We consider $25$ users and mean opinion scores are shown in \cref{tab:user_study}.

\subsection{Object Manipulation}

\begin{figure*}
    \centering
    \begin{subfigure}[b]{0.24\textwidth}
         \centering
         \includegraphics[width=0.92\linewidth]{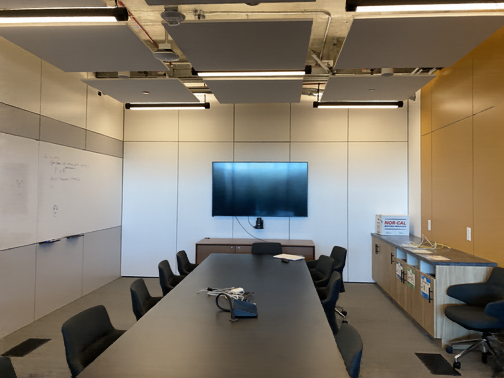}
     \end{subfigure}
     \begin{subfigure}[b]{0.24\textwidth}
         \centering
         \includegraphics[width=0.92\linewidth]{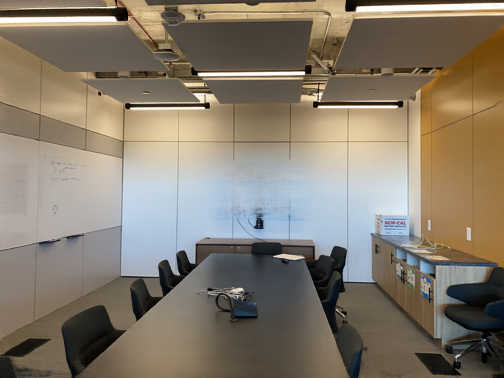}
     \end{subfigure}
     \begin{subfigure}[b]{0.24\textwidth}
         \centering
         \includegraphics[width=0.92\linewidth]{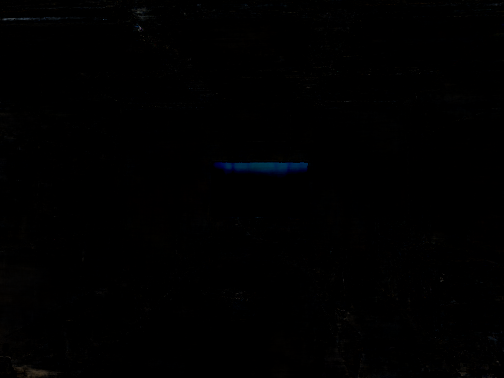}
     \end{subfigure} 
     \begin{subfigure}[b]{0.24\textwidth}
         \centering
         \includegraphics[width=0.92\linewidth]{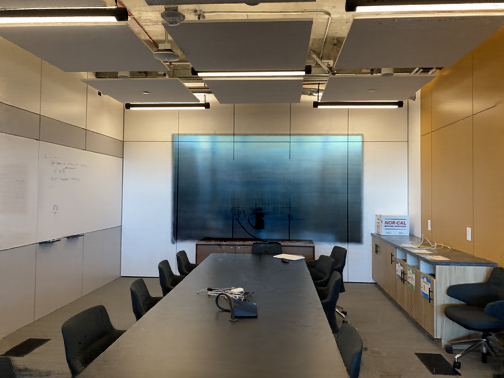}
     \end{subfigure} \\
     
    \begin{subfigure}[b]{0.24\textwidth}
         \centering
         \includegraphics[width=0.92\linewidth]{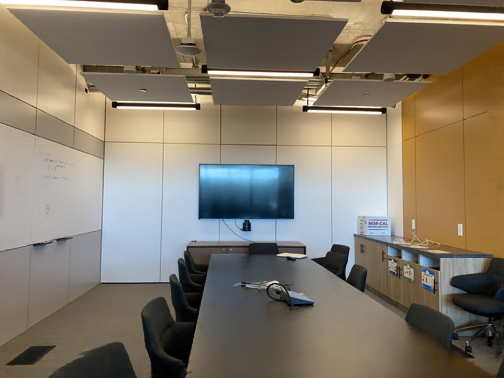}
         \caption{Original Scene}
     \end{subfigure}
     \begin{subfigure}[b]{0.24\textwidth}
         \centering
         \includegraphics[width=0.92\linewidth]{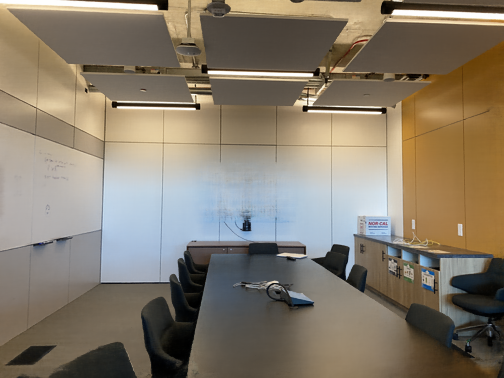}
         \caption{Background}
     \end{subfigure}
     \begin{subfigure}[b]{0.24\textwidth}
         \centering
         \includegraphics[width=0.92\linewidth]{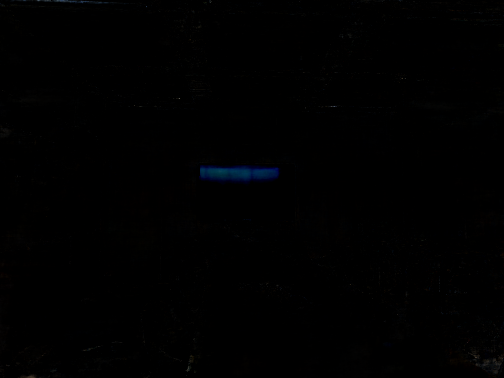}
         \caption{Foreground Obj.}
     \end{subfigure} 
     \begin{subfigure}[b]{0.24\textwidth}
         \centering
         \includegraphics[width=0.92\linewidth]{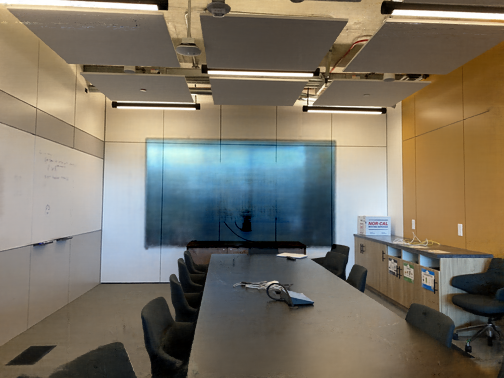}
         \caption{New Scene}
     \end{subfigure}
    \caption{\textbf{Foreground object transformation}. Our method makes plausible predictions in occluded regions (behind the TV) by understanding the correlated effects from the rest of the scene, such as the light reflections in the TV screen, which are not visible in the foreground. After scaling the foreground object and placing it back into the scene, the correlated effects are introduced again, resulting in photo-realistic and view consistent light reflections on the TV screen.}
    \label{fig:tv_scaling}
    \vspace{-0.3cm}
\end{figure*}

\paragraph{Foreground Transformation.}
We consider the ability to scale the foreground object and place the rescaled object back into the scene
by changing the focal length used to generate the rays of the foreground object, and then volumetrically adding it back into our background volume. \cref{fig:tv_scaling} shows an example where the disentangled TV is twice as large. We note that other transformations such as translation and rotation are possible in a similar manner. \cref{fig:tv_scaling} highlights several properties of our volumetric disentanglement volume. First, the network is able to ``hallucinate" how a plausible background looks in regions occluded across all views (\textit{e.g.} behind the TV). It does this based on correlated effects from the rest of the scene. A second property is that it can disentanglement correlated effects such as the reflections on the TV screen, which is evident from the almost completely black TV in the foreground scene. Lastly, these correlated effects result in consistent and photo-realistic reflections, when we place the rescaled TV back into the scene. These reflections are consistent across views.

\begin{figure*}
\centering
\begin{tabular}{ccccc}
         \includegraphics[width=0.182\linewidth] {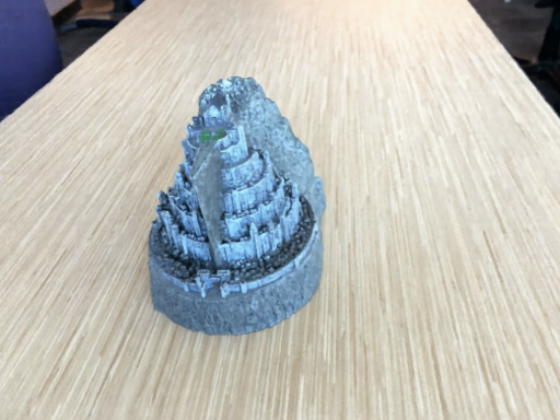} &
         \includegraphics[width=0.182\linewidth]{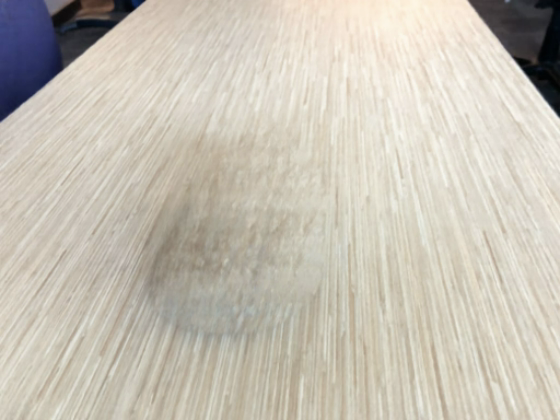} &
         \includegraphics[width=0.182\linewidth]{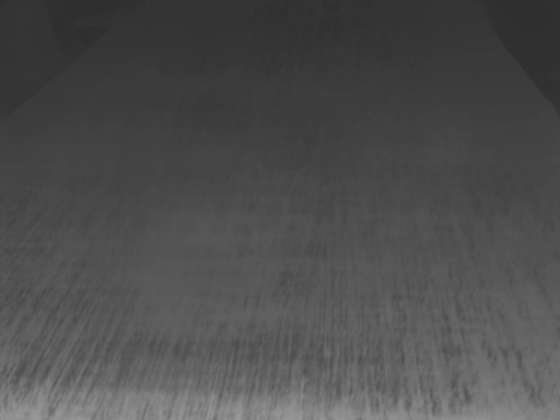} &
         \includegraphics[width=0.182\linewidth]{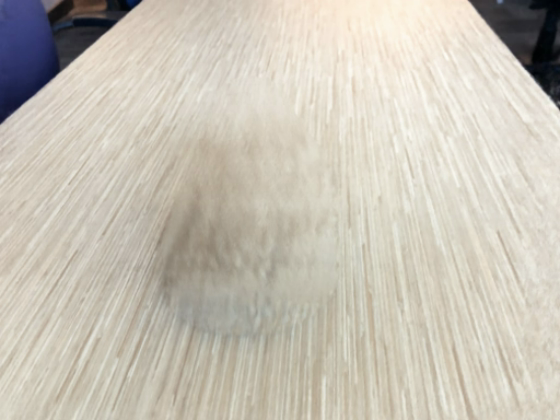} &
         \includegraphics[width=0.182\linewidth]{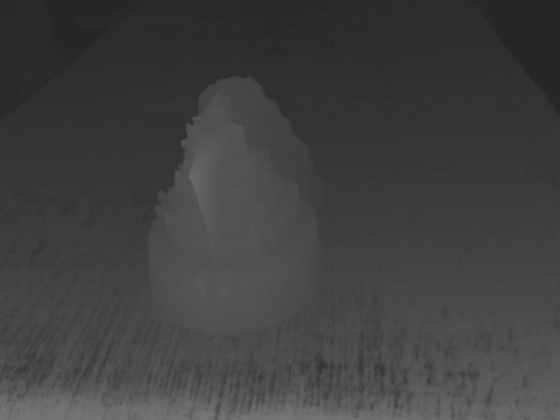} \\ 
         \includegraphics[width=0.182\linewidth]{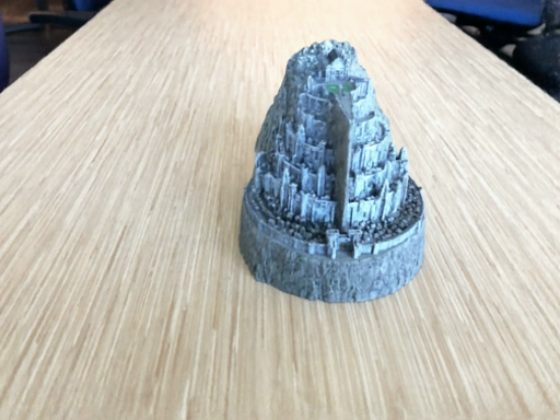} &
         \includegraphics[width=0.182\linewidth]{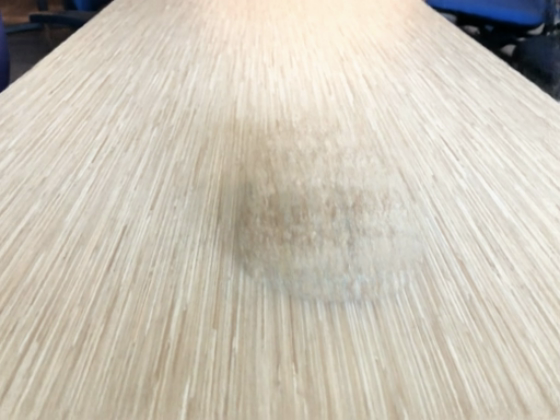} &
         \includegraphics[width=0.182\linewidth]{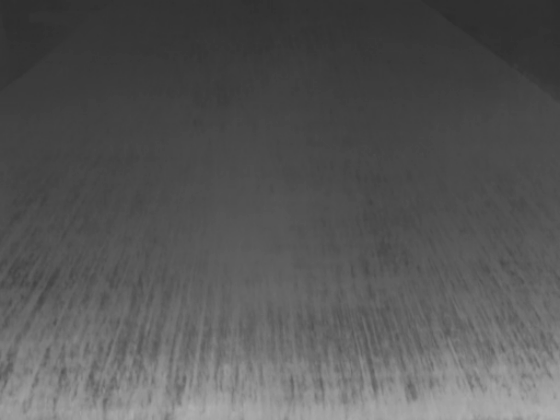} &
         \includegraphics[width=0.182\linewidth]{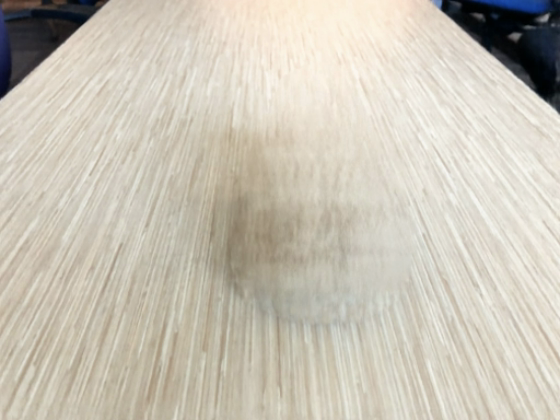} &
         \includegraphics[width=0.182\linewidth]{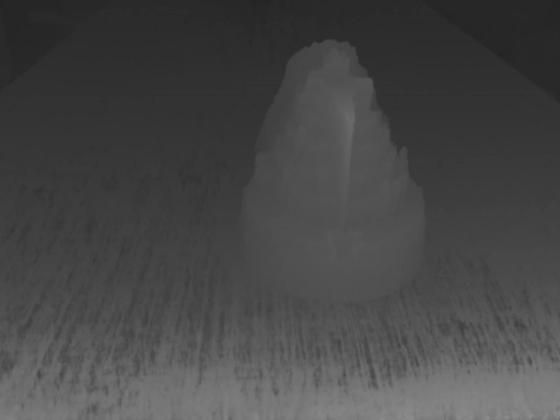} \\ 
         (a) &(b) &(c) &(d) & (e) \\
         \vspace{-0.5cm}
     \end{tabular}
     \caption{\textbf{Object camouflage for two different random views of a fortress scene.} (a) original scene, (b) background scene,  (c) disparity map of the background scene, (d) camouflaged scene, (e) disparity map of camouflaged scene.  }

        \label{fig:camouflage}
        \vspace{0.1cm}

         \includegraphics[width=0.97\linewidth]{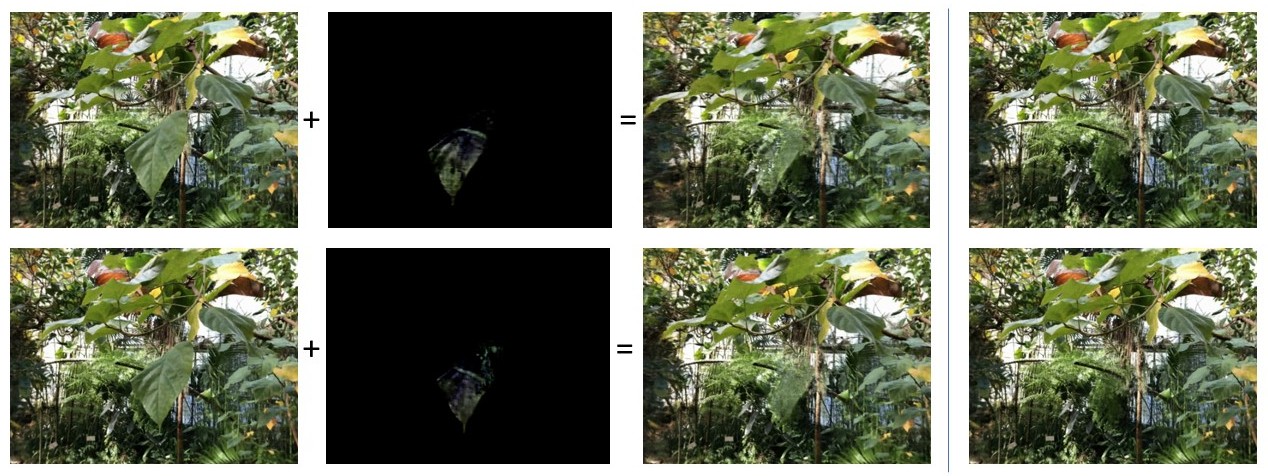}
         
         ~~~~~~~~~~~~  (a) ~~~~~~~~~~~~~~~~~~~~~  (b) ~~~~~~~~~~~~~~~~~~~~~~  (c) ~~~~~~~~~~~~~~~~~~~~~~~~  (d) \\
         \vspace{-0.1cm}

        \caption{\textbf{Non-negative object inpainting for two views for a scene of leaves.} Given the full scene (a), a residual scene is added (b) resulting in scene (c), with the aim of being close to the background without the leaf (d).  }
        \label{fig:staypositive}
\begin{center}
    \includegraphics[width=0.97\textwidth]{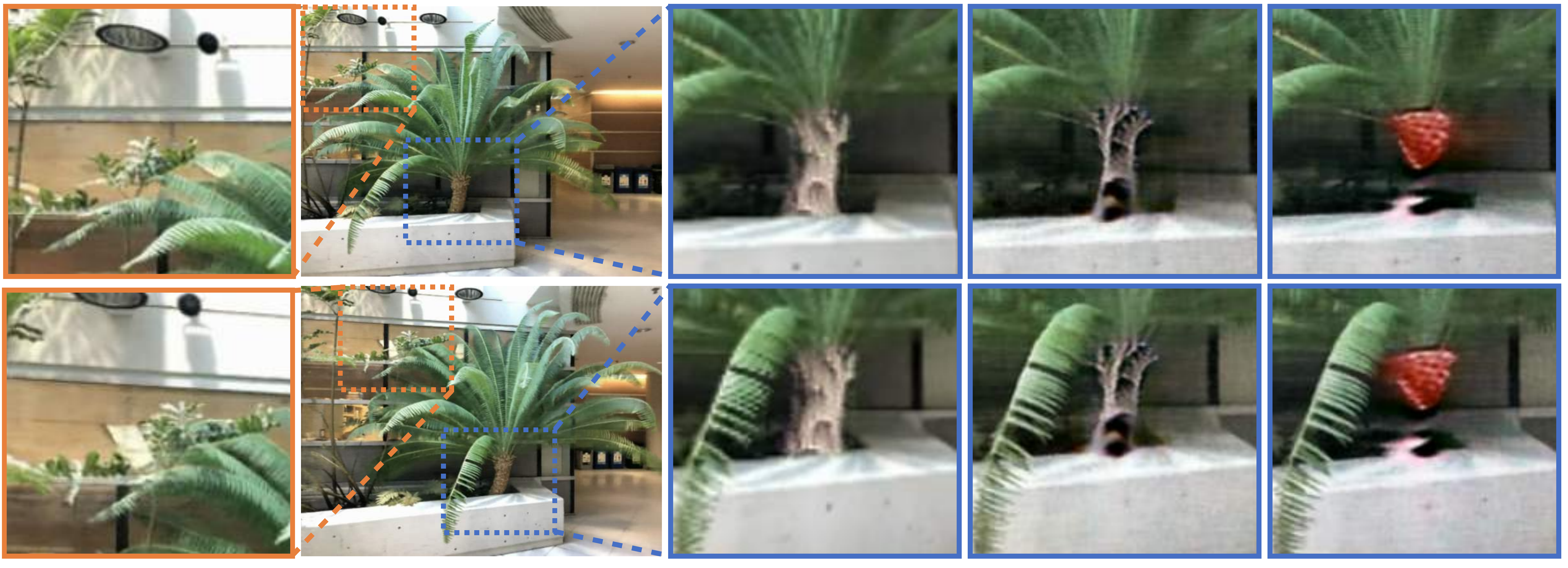} 
    
    ~~ (a) ~~~~~~~~~~~~~~~~~~  (b) ~~~~~~~~~~~~~~~~~~  (c) ~~~~~~~~~~~~~~~  (d) ~~~~~~~~~~~~~~~  (e) ~~~~\\
    
    \caption{\textbf{3D Object Manipulation.} Insets of the disentangled (a) window mullion and manipulated (c)-(e) tree trunk in the original scene (b). Note how the window mullion is removed without removing the leaf of the fern that occludes it from the first view. The query text to manipulate the trunks are (c) \textit{Old tree}, (d) \textit{Aspen tree}, and (e) \textit{Strawberry}. The manipulated objects are view-consistent.}
    \label{fig:fern_figure}
\end{center}
\end{figure*}

\paragraph{Object Camouflage.}
Another manipulation of interest is of camouflaging an object~\cite{owens2014camouflaging,guo2022ganmouflage}, \textit{i.e.} only change the texture of the object and not its shape.
\cref{fig:camouflage} illustrates examples of camouflaging with fixed depth, but free texture changes. While the depth of the camouflaged object and that of the foreground object match, the appearance of the camouflaged object is that of the background. 

\paragraph{Non-Negative Inpainting.}
In optical see-through AR, one might also wish to camouflage objects~\cite{luo2021stay} or inpaint them. However, in see-through AR one can only add light. 
\cref{fig:staypositive} shows how adding light can make the appearance of camouflage in a 3D consistent manner.

\paragraph{3D Object Manipulation.}
We now demonstrate how our disentanglement can be used for 3D object manipulation. \cref{fig:fern_figure} shows two views of a fern. We have disentangled both the window mullion in the upper left corner and the tree trunk from the rest of the scene. Even though the window mullion is occluded in the first view, and thus our 2D mask is masking the occluding leaf in front of the window mullion, this occluding object is not part of the disentangled window mullion object. 
The 3D manipulations are shown in (c)-(e) in \cref{fig:fern_figure}. Notice how the manipulated 3D objects are semantically consistent across views. For the strawberry manipulation in (e), note how part of the tree trunk was camouflaged to more closely resemble the shape of a strawberry. We compare to 2D text-based inpainting methods of GLIDE~\cite{nichol2021glide} and Blended Diffusion~\cite{avrahami2021blended}, where we follow the same procedure as in \cref{sec:object_disent}. We consider a similar user as detailed in \cref{sec:object_disent}, where Q1 is modified to: ``How well was the object semantically manipulated according to the target text prompt?'' For the user study we consider the fern scene of \cref{fig:fern_figure}, for the text prompts of ``strawberry'' and ``old tree''.

\begin{figure*}[t]
\centering
\includegraphics[width=0.915\linewidth]{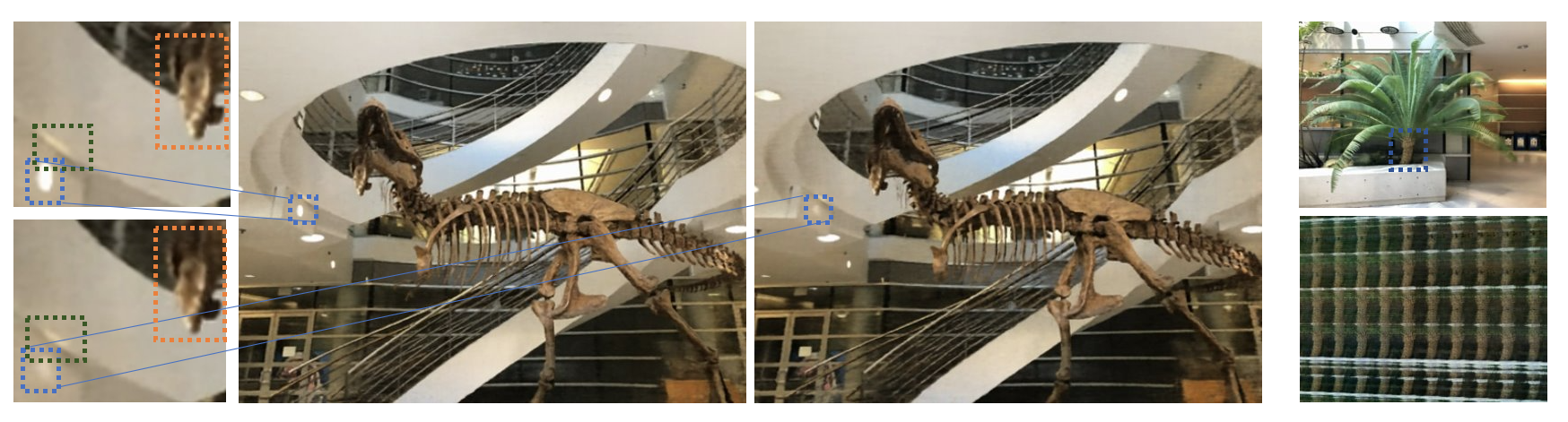} \\ 
\vspace{-0.17cm}
~~~~~~~~~~~~~~~~~~~~~~~~~~~~ (a) ~~~~~~~~~~~~~~~~~~~~~~~~~~~~~~~~~~~~~~~~~~~~~~~~~~~~(b) \\
\vspace{0.1cm}
\begin{tabular}{cccc}
         \includegraphics[width=0.2172\linewidth] {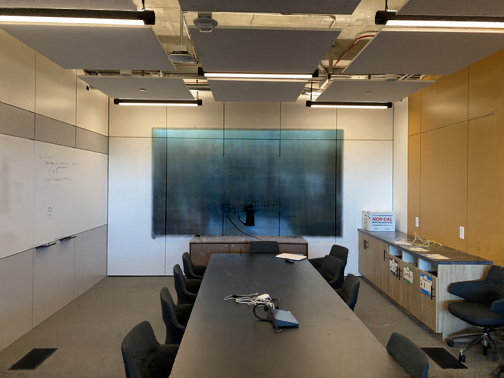} &
         \includegraphics[width=0.2172\linewidth]{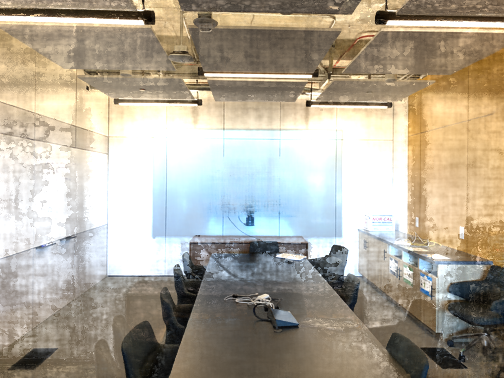} &
         \includegraphics[width=0.2172\linewidth]{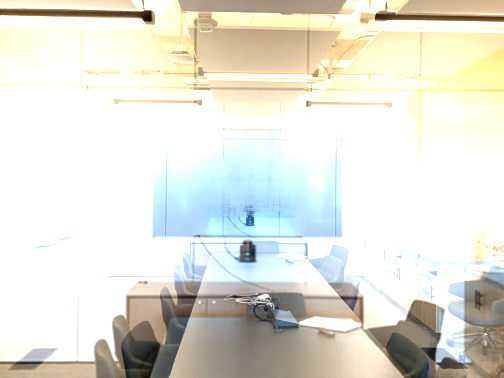} &
         \includegraphics[width=0.2172\linewidth]{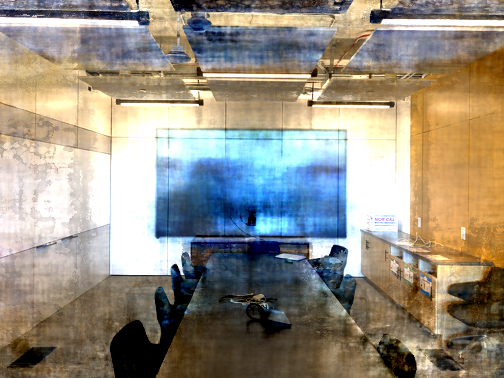} \\
         (c1 - Ours) & (c2)& (c3) & (c4)\\ 
     \end{tabular}
      
\begin{tabular}{cccc}
         \includegraphics[width=0.2172\linewidth] {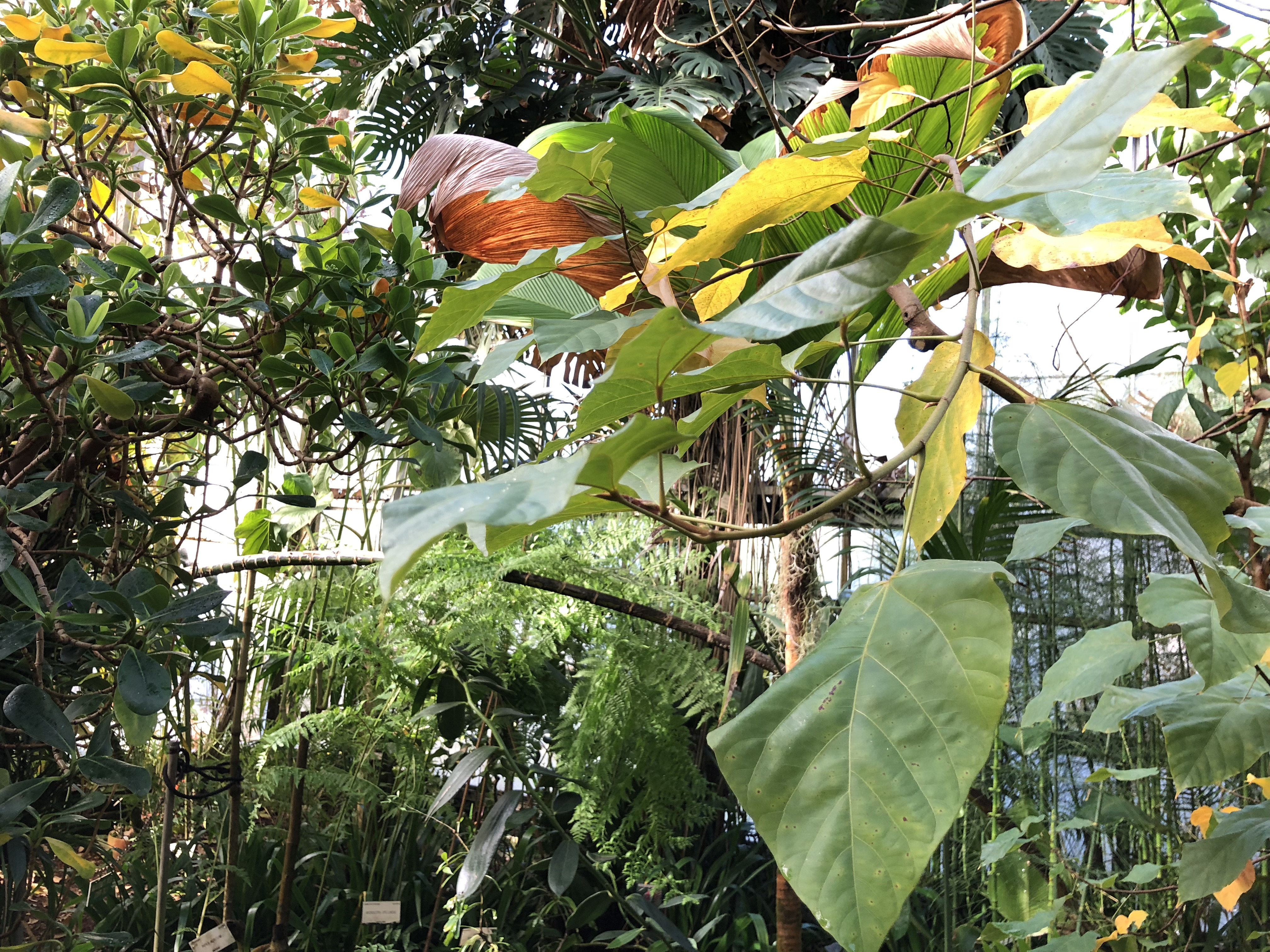} &
         \includegraphics[width=0.2172\linewidth]{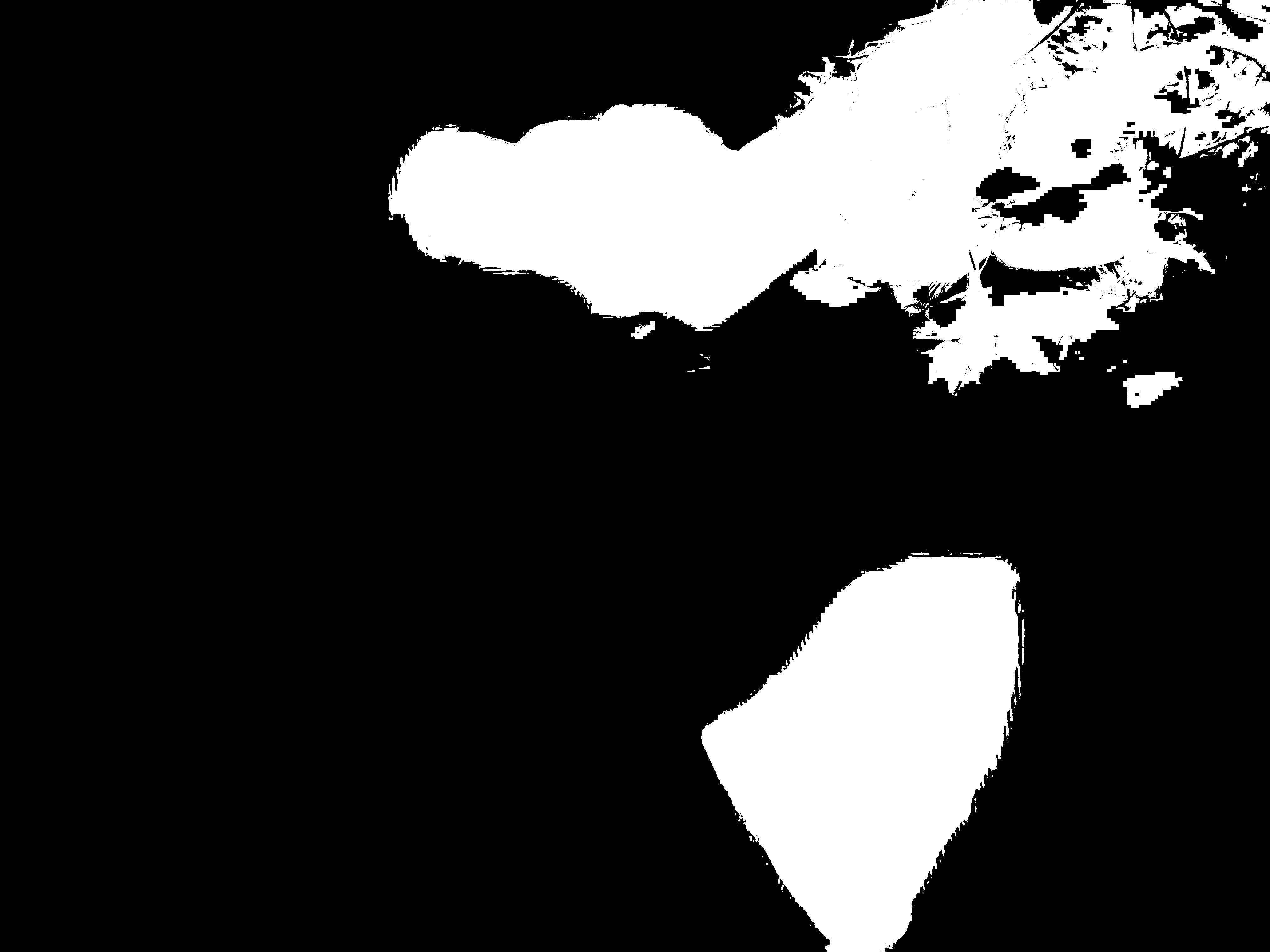} &
         \includegraphics[width=0.2172\linewidth]{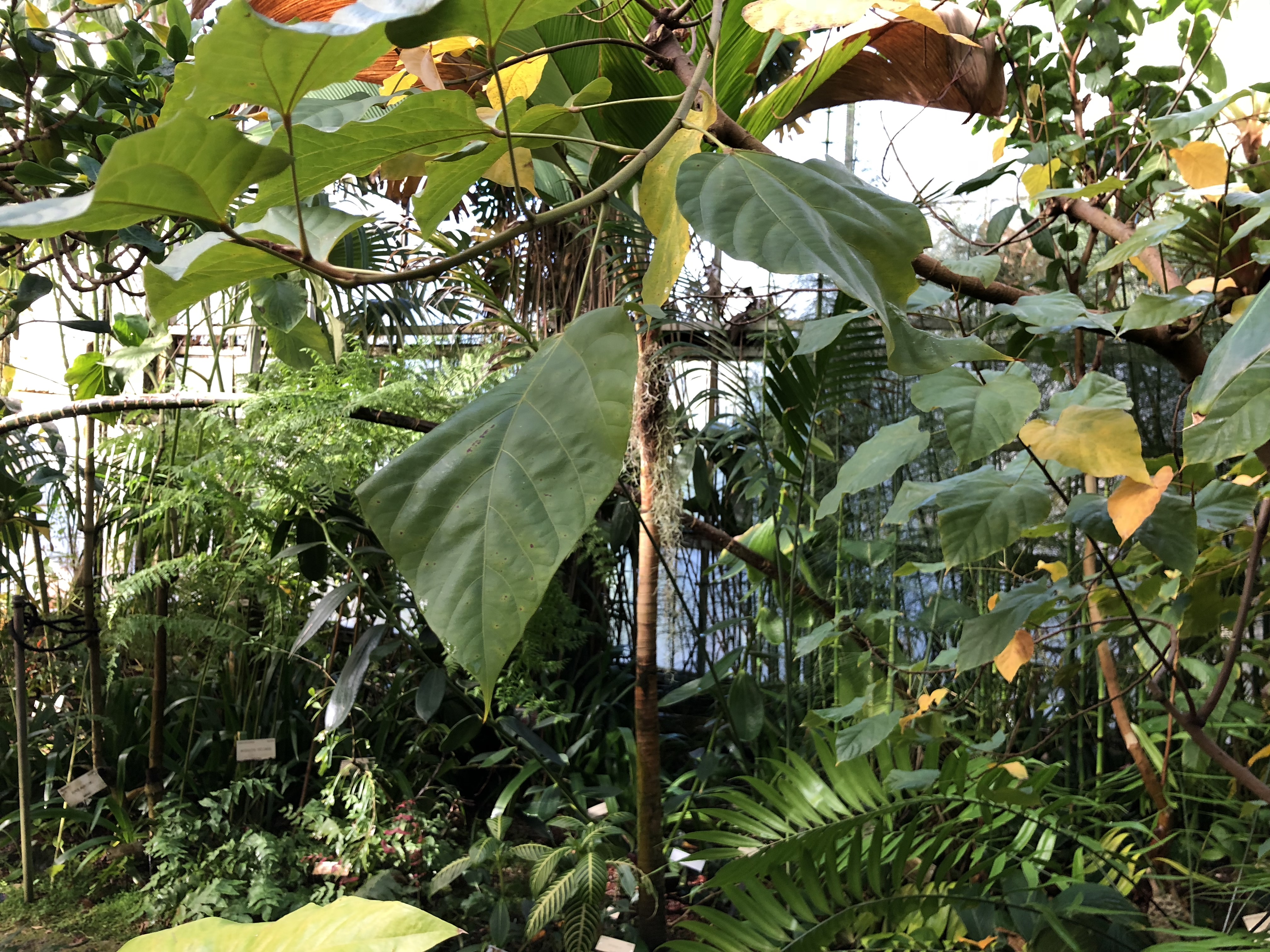} &
         \includegraphics[width=0.2172\linewidth]{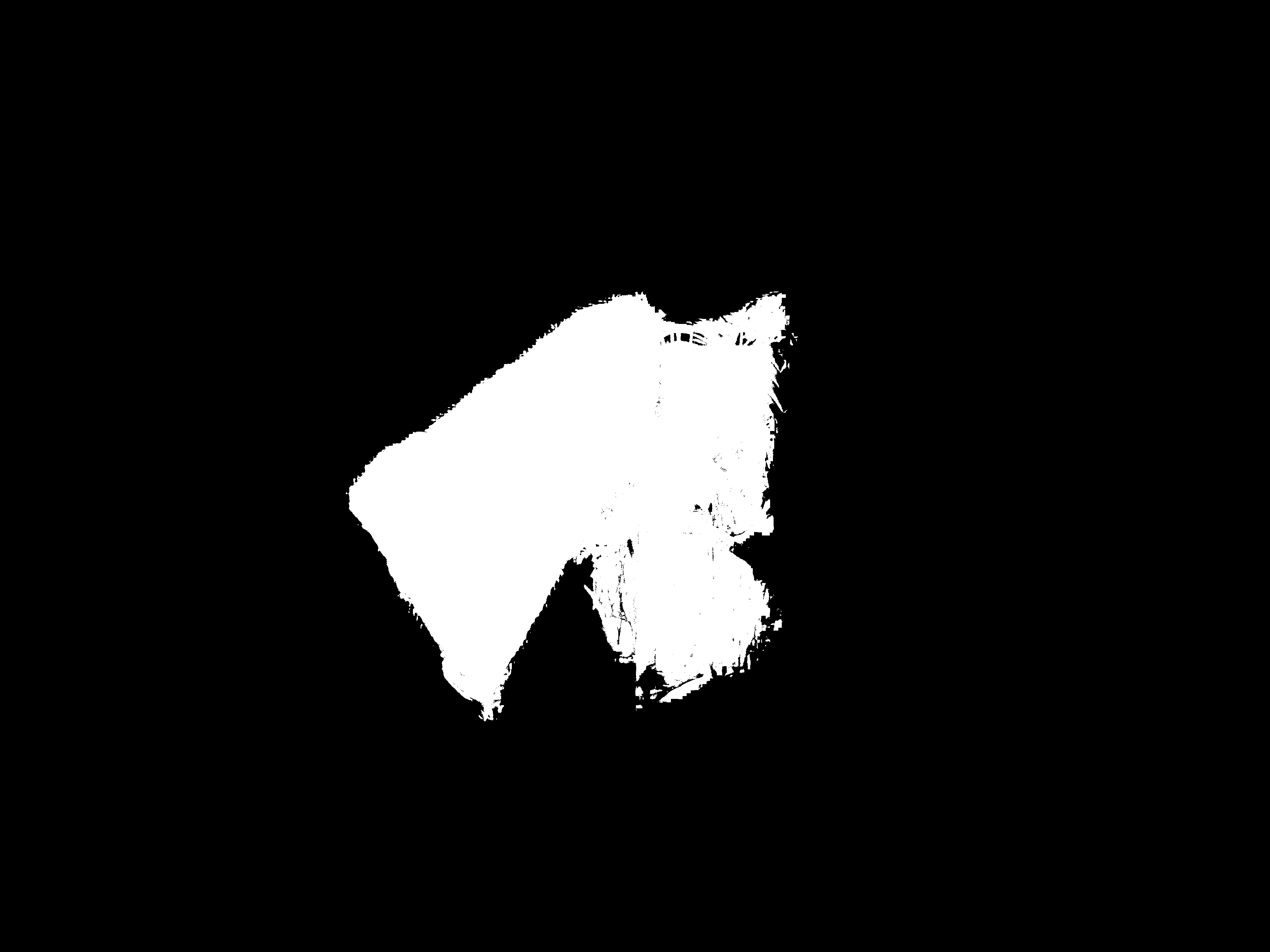} \\
     \end{tabular}
     
     (d) 
     \vspace{-0.2cm}
\caption{ 
(a) \textit{Failure to completely remove a light source.} The original light source is shown in blue in the middle image and for the background, using our method, on the right. In orange and green are regions affected by the light source, resulting in the failure to completely remove it. 
(b) \textit{Illustration of the result of training a neural radiance field on the masked foreground region.} (c1-c4) \textit{Ablation for composition}. Alternatives to the composition shown in \cref{eq:composition} for foreground object translation (\cref{fig:tv_scaling}). 
(d) \textit{Robustness to noisy 2D masks.}  Our method can handle noisy 2D masked obtained automatically. 
}
\label{fig:discussion}
\vspace{-0.3cm}
\end{figure*}

\subsection{Discussion and Limitations}
\label{sec:discussion}

Our work has some limitations. 
When light from the background affects the foreground object, we correctly disentangle the illuminations on the object. However, when the object is a light source, we cannot completely disentangle the object as seen in \cref{fig:discussion}(a). We also note that, while our work can handle noisy masks, we require masks for all training views. We leave the task of reducing the number of masks for future work. 

Another limitation is with respect to the semantic manipulation of foreground objects. We found that manipulating too large objects results in an under-constrained optimization because the signal provided by CLIP is not sufficient. 
For smaller objects, the background from multiple views provides a much-needed context for CLIP to provide a useful signal for the optimization. 
We note that our work is orthogonal to recent speed and generalization extensions of NeRF that could combined with our method. The number of 2D masks required by our method is also upper bounded by the number of training views and so methods such as DietNeRF~\cite{jain2021putting} could be combined with our method to reduce the number of 2D masks required. We leave such combinations to future work.

In \cref{fig:discussion}(b), we illustrate the necessity of extracting the foreground volume from the background and the full volume, rather than directly learning the foreground volume. We tried to train a neural radiance field to reconstruct the foreground volume directly, which resulted in an under-constrained optimization. 

In Fig.~\ref{fig:discussion} (c1 to c4), we show, for the task of foreground object translation (\cref{fig:tv_scaling}), alternatives to the recombining method of \cref{eq:composition}, with (c2) ${c'_{full}}^{i_r}$ instead of ${c'_{fg}}^{i_r}$, (c3) ${w'_{full}}^{i_r}$ instead of ${w'_{fg}}^{i_r}$, (c4)
$c^{c}_r = \sum_{i=1}^N ({w'_{bg}}^{i_r} + {w'_{fg}}^{i_r}) \cdot ({c'_{bg}}^{i_r} +  {c'_{fg}}^{i_r})$. 

Lastly, we note that our method can handle noisy annotations of the foreground. In \cref{fig:discussion}(d), we demonstrate the masks used for the leaves scene, which were extracted using an off-the-shelf segmentation algorithm.

\section{Conclusion}

In this work, we presented a framework for volumetric disentanglement of foreground objects from a background scene. 
The disentangled foreground object is obtained by volumetrically subtracting a learned volume representation of the background with one from the entire scene. The foreground-background disentanglement adheres to object occlusions and background effects such as illumination and reflections. We established that our disentanglement facilitates separate control of color, depth, and transformations for both the foreground and background objects. This enables a wide range of applications, of which we have demonstrated those of foreground transformations, object camouflage, non-negative generation, and 3D object manipulation. We hope that our volumetric disentanglement framework can inspire designers and artists to better express their creativity across a wide range of augmented reality applications.

\section*{Acknowledgement}

This research was supported by the Pioneer Centre for AI, DNRF grant number P1. We would like to thank Zekun Hao for the helpful discussions.

\clearpage
%
%
\bibliographystyle{splncs04}
\bibliography{egbib}
\end{document}